\pdfoutput=1

\documentclass[11pt]{article}

\PassOptionsToPackage{table,xcdraw}{xcolor}
\usepackage[]{acl}

\usepackage{times}
\usepackage{latexsym}

\usepackage[T1]{fontenc}

\usepackage[utf8]{inputenc}

\usepackage{microtype}

\usepackage{inconsolata}

\usepackage[most]{tcolorbox}
\usepackage{graphicx}
\usepackage{caption}
\usepackage{algorithm}
\usepackage{algorithmic}
\usepackage{multirow}
\usepackage{amsfonts,amsmath,amssymb}
\usepackage{subcaption}
\usepackage[capitalize]{cleveref}
\usepackage{enumitem}
\usepackage{pifont}
\usepackage{newfloat}
\usepackage{listings}
\usepackage{soul}
\usepackage{tikz}
\usepackage{tabularx}
\usepackage{booktabs}
\usepackage{wrapfig}
\usepackage{array}
\usepackage{minitoc}
\usepackage{microtype}

\newcommand{\algodataset}{\textsc{AlgoPuzzleVQA}}
\newcommand{\abstractdataset}{\textsc{PuzzleVQA}}

\newcommand{\greyrule}[2]{%
    \arrayrulecolor{black!30}%
    \cmidrule(lr){#1-#2}%
    \arrayrulecolor{black}%
}

\usepackage{pifont}

\definecolor{customgreen}{rgb}{0.325, 0.675, 0.196}

\definecolor{NotSureBlue}{rgb}{0.396, 0.443, 0.627}

\usepackage{fvextra} 
\DefineVerbatimEnvironment{MyVerbatim}{Verbatim}{
  breaklines=true,   
  breakanywhere=true,
  breaksymbolleft={}
}

\definecolor{curiosity}{rgb}{0.58, 0.0, 0.83} 

\title{\textbf{\textit{\textcolor{curiosity}{The Jumping Reasoning Curve?}}}\\ 
Tracking the Evolution of Reasoning Performance in GPT-[n] and o-[n] Models on Multimodal Puzzles}

\author{Vernon Y.H. Toh$^1$, Yew Ken Chia$^1$, Deepanway Ghosal$^1$, Soujanya Poria$^1$ \\\\
$^1$ Singapore University of Technology and Design
}

\usepackage{colortbl}

\usetikzlibrary{shadows.blur, fadings}

\definecolor{headerblue}{RGB}{31,74,126}
\definecolor{headergray}{RGB}{80,80,80}
\definecolor{rowgray}{RGB}{245,245,245}

\newcolumntype{C}{>{\centering\arraybackslash}p{2cm}}

\begin{document}

\maketitle

\begin{abstract}

The releases of OpenAI's o-[n] series, such as o1, o3, and o4-mini, mark a significant paradigm shift in Large Language Models towards advanced reasoning capabilities.
Notably, models like o3 have demonstrated strong performance on benchmarks like the Abstraction and Reasoning Corpus for Artificial General Intelligence (ARC-AGI).
However, this benchmark is limited to symbolic patterns, whereas humans often perceive and reason about multimodal scenarios involving both vision and language data.
Thus, there is an urgent need to investigate advanced reasoning capabilities in multimodal tasks.
To this end, we track the evolution of the GPT-[n] and o-[n] series models (including o1, o3, and o4-mini) on challenging multimodal puzzles from \textsc{PuzzleVQA} and \textsc{AlgoPuzzleVQA}, which demand fine-grained visual perception.
Our results reveal that o-[n] series, particularly later iterations like o3 and o4-mini, significantly outperform the GPT-[n] series and show strong scalability in multimodal reasoning.
\textit{Nonetheless, despite these substantial advancements and the superior capabilities demonstrated by the o-[n] series, our findings highlight that even these leading models face persistent challenges. Difficulties are particularly evident in tasks requiring precise visual perception, robust compositional reasoning across multiple visual attributes, and solving complex algorithmic or highly combinatorial puzzles, indicating critical areas for future AGI development.}
We plan to continuously track new models in the series and update our results in this paper accordingly.
All resources used in this evaluation are openly available at \url{https://github.com/declare-lab/LLM-PuzzleTest}.

\end{abstract}

\begin{figure}[ht]
    \centering
    \includegraphics[width=0.70\linewidth]{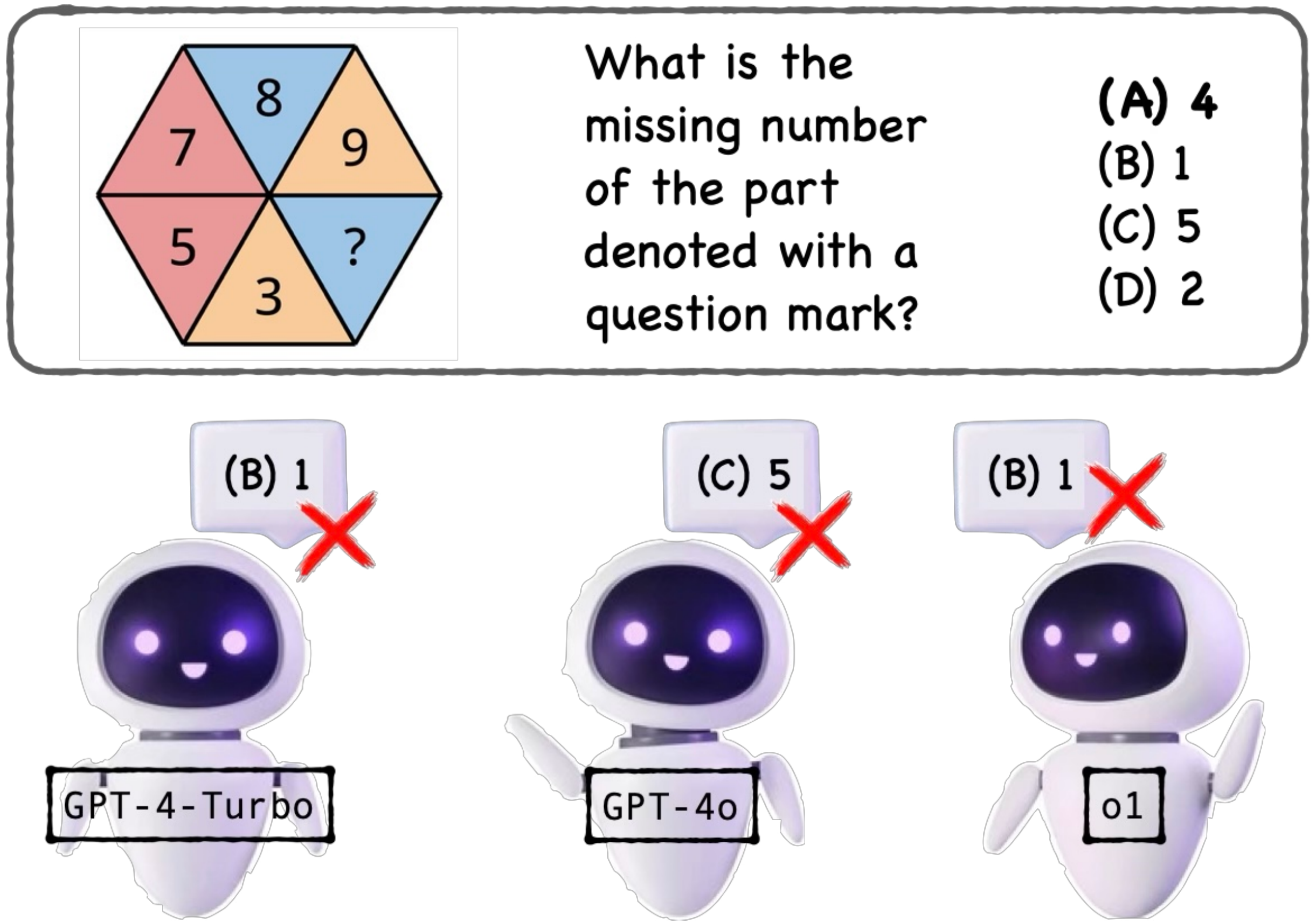}
    \caption{Puzzle from \abstractdataset{}. GPT-4-Turbo, GPT-4o, and o1 all got the puzzle incorrect.}
    \label{fig:intro_case}
\end{figure}

\begin{figure*}[ht]
    \centering
    \includegraphics[width=0.99\textwidth]{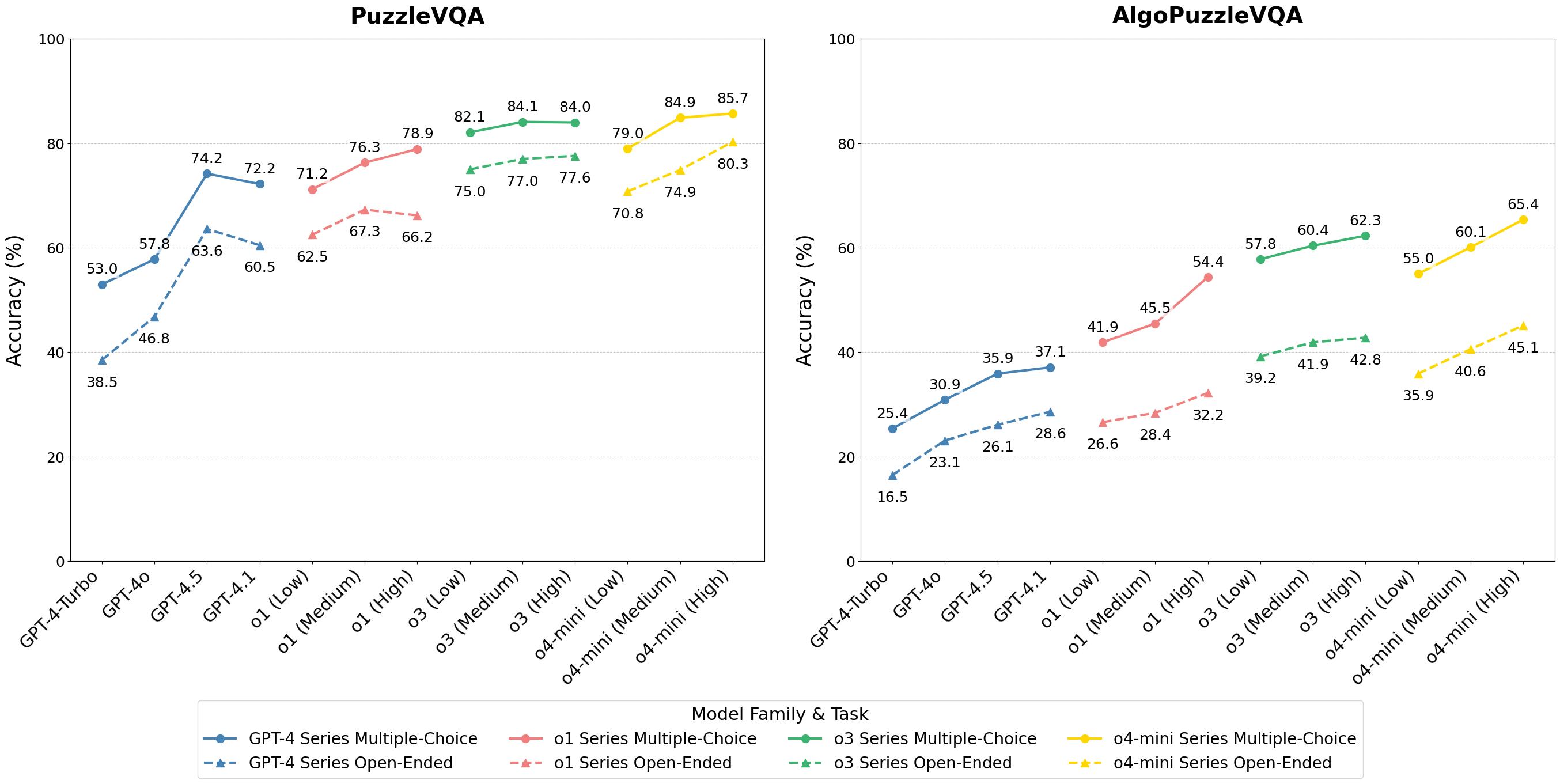}
    \caption{The performance of GPT-[n] and o-[n] series models on \abstractdataset{} and \algodataset{}, illustrating how multimodal reasoning evolves over time with model releases.}
    \label{fig:visual_trends}
\end{figure*}

\section{Introduction}
Recent advances in large language models (LLMs) have demonstrated impressive capabilities in language understanding and generation, as seen in OpenAI's GPT-[n] series of models \citep{gpt3}.
Yet, true artificial general intelligence (AGI) requires robust reasoning abilities across different modalities \citep{Fei2021TowardsAG}.
Simply excelling in text-based tasks is insufficient for agents intended to understand and interact with the rich, multifaceted physical world humans navigate daily.
For instance, while models such as OpenAI's new o-[n] series demonstrate significant performance improvements, sometimes characterized as a ``jumping reasoning curve'' on specific benchmarks like the Abstraction and Reasoning Corpus for Artificial General Intelligence (ARC-AGI) \citep{chollet2019measureintelligence}, the landscape of multimodal reasoning is complex.
The current evaluations in this area often focus on symbolic patterns, whereas humans frequently reason over complex data involving both vision and language.
Thus, the ability to perceive, understand, and reason about multimodal inputs remains a crucial component of human-like intelligence, deserving urgent investigation.

To this end, puzzles often serve as effective measures of cognitive abilities such as pattern recognition and step-by-step reasoning.
Notably, such measures typically do not require specific domain knowledge, allowing individuals from diverse backgrounds to engage with them. One prominent example is Raven's Progressive Matrices \citep{raven-ebe0e932-0e0b-3cbd-9c98-9b94f8c2f1dc}, a non-verbal assessment tool designed to evaluate abstract reasoning and fluid intelligence. In this test, participants are presented with abstract patterns containing a missing element and must identify the correct piece to complete the pattern.

Thus, inspired by abstract puzzles as measures of intelligence, recent multimodal benchmarks have enabled systematic evaluation across specific cognitive abilities, including visual perception, inductive reasoning, deductive reasoning, and algorithmic problem solving \citep{chia2024puzzlevqadiagnosingmultimodalreasoning, ghosal2024languagemodelspuzzleprodigies}.
These newer benchmarks go beyond simple object recognition or image captioning, probing deeper into how models integrate information across modalities to make inferences or devise multi-step solutions.
Compared to previous measures, they require general understanding of spatial relationships, pattern recognition, and reasoning across visual and language elements, thus providing a more holistic measure of artificial general intelligence.
Our research addresses several key questions:
(1) How do current state-of-the-art models, specifically the GPT-[n] and o-[n] series, perform on diverse visual reasoning tasks?
(2) What types of pattern recognition, compositional reasoning, and algorithmic problem-solving are particularly challenging for these models?
(3) How can we systematically evaluate and compare different models' multimodal reasoning capabilities?

In our evaluation, we assess the performance of GPT-[n] and o-[n] models on abstract multimodal puzzles from \textsc{PuzzleVQA}, which primarily test abstract reasoning, and on \textsc{AlgoPuzzleVQA}, which requires more complex algorithmic approaches. To ensure a comprehensive evaluation, we present the puzzles in both multiple-choice and open-ended question answering formats.

Our findings reveal that while the o-[n] series demonstrates superior performance and scalability, significantly outperforming GPT-[n] models across most tasks, even these advanced models encounter substantial difficulties, particularly with complex algorithmic reasoning and fine-grained visual perception.
The performance gains, while notable within the o-[n] series, highlight specific bottlenecks rather than uniform leaps across all reasoning types, especially when compared to the broader challenges posed by our benchmarks. For example, GPT-[n] models show iterative improvements but consistently lag, struggling with visual perception and more complex compositional tasks.
This suggests that merely scaling existing architectures may not be sufficient to overcome certain fundamental hurdles in perception and logic.
This differentiation in capabilities and the persistent challenges, even for leading models, underscore the substantial gap between current artificial intelligence and human-like reasoning abilities.
As models continue to rapidly advance and scale, as suggested by trends like those in \Cref{fig:visual_trends}, this benchmark and the detailed analysis of performance tiers and bottlenecks will serve as a critical indicator of progress toward more robust and generalized artificial intelligence.

\section{The Path to AGI: Why Solving Puzzles and Recognizing Abstract Patterns are Key Milestones}

There isn't a single universally accepted definition of Artificial General Intelligence (AGI).
However, following \cite{xu2024meantagidefinitionartificial}, AGI refers to a system capable of adapting to novel open environments using limited computational resources, guided by specific principles.
In contrast to narrow AI, which excels in specific tasks, AGI aims to replicate the broad cognitive abilities of humans, enabling it to perform any intellectual task that a human can \citep{latif2024agiartificialgeneralintelligence}.
Achieving AGI involves creating systems that can learn and reason under conditions of insufficient knowledge and resources, continuously adapting to new tasks and environments \cite{johansson2024machinepsychologyintegratingoperant}.

Solving puzzles is a significant milestone in the path to AGI because it requires the application of core cognitive skills such as abstract reasoning, problem-solving, and pattern recognition. These skills are essential for a model to generalize knowledge and adapt to new, unseen tasks. The process of puzzle-solving involves several cognitive functions central to human intelligence, including the ability to understand and manipulate abstract concepts, recognize patterns, and apply logical reasoning to arrive at a solution. By evaluating these models on puzzles, we can assess and enhance these cognitive functions within the AI, bringing it closer to human-like intelligence \cite{estermann2024puzzlesbenchmarkneuralalgorithmic}.

\citet{wüst2024bongardwonderlandvisualpuzzles} discuss how current models occasionally succeed in identifying discriminative concepts but often fail to understand and reason about visual concepts, indicating a significant limitation in their reasoning abilities. 
Similarly, \citet{park2023unravelingarcpuzzlemimicking} explore how models approach the Abstraction and Reasoning Corpus (ARC), revealing both their strengths and limitations in abstract reasoning.

The release of o-[n] models highlights the relationship between computational resources and performance.
While increased computational power can enhance performance, true intelligence also requires efficient learning algorithms and the ability to generalize from limited data \cite{sastry2024computingpowergovernanceartificial}. 
Simply scaling up resources does not guarantee the emergence of AGI; developing architectures that can learn abstract concepts and apply them across various domains is crucial \cite{mumuni2025largelanguagemodelsartificial}.

\newpage
\subsection{Performance on Existing Puzzle Benchmarks}

The performance of GPT-[n] and O-[n] models across various puzzle benchmarks reveals both their strengths and limitations in reasoning and multimodal tasks.

\citet{giadikiaroglou2024puzzlesolvingusingreasoning} highlights that, despite advancements, current models often struggle with complex textual rule-based and rule-less puzzles requiring advanced logical inference, revealing a significant gap between their capabilities and human-like reasoning.

On the \textsc{NPR Sunday Puzzle} benchmark \cite{anderson2025phdknowledgerequiredreasoning}, o1 achieved a 59\% accuracy rate, outperforming models like R1, which demonstrates its proficiency in general reasoning. However, GPT-4o struggled with deep abductive reasoning tasks, scoring only 38\% on the \textsc{True Detective} benchmark \cite{del2023truedetectivedeepabductive}.

In the \textsc{REBUS} benchmark \cite{gritsevskiy2024rebusrobustevaluationbenchmark}, GPT-4o performed well with a 42\% accuracy rate, but faced difficulties with more complex puzzles, achieving just 7\% on the hardest ones. The o1 model also showed limitations on the \textsc{EnigmaEval} benchmark \cite{wang2025enigmaevalbenchmarklongmultimodal}, with only 7\% accuracy on the normal split and 0\% on the hard split, highlighting its struggle with unstructured and lateral reasoning tasks.

GPT-4V demonstrated moderate success on the \abstractdataset{} \cite{chia2024puzzlevqadiagnosingmultimodalreasoning} benchmark with a 46\% accuracy rate, showcasing its ability in abstract pattern recognition. However, it scored only 30.3\% on the \algodataset{} \cite{ghosal2024languagemodelspuzzleprodigies}, revealing challenges in solving complex algorithmic puzzles that require both visual and algorithmic reasoning.

Unlike other puzzle benchmarks, \abstractdataset{} tests a model’s understanding of basic concepts such as colors, shapes, sizes, and spatial relationships. This sets it apart from datasets that focus purely on text or visual recognition. \algodataset{} increases the complexity by introducing algorithmic reasoning, requiring models to solve puzzles that combine visual comprehension with intricate algorithmic thinking. Both datasets allow for deeper analysis, as they are created with predefined ontologies, making it easier to identify reasoning gaps. These qualities make \abstractdataset{} and \algodataset{} the ideal benchmarks for our study, providing a comprehensive foundation for our analysis.

\begin{figure*}[ht]
    \centering
    \includegraphics[width=0.76\linewidth]{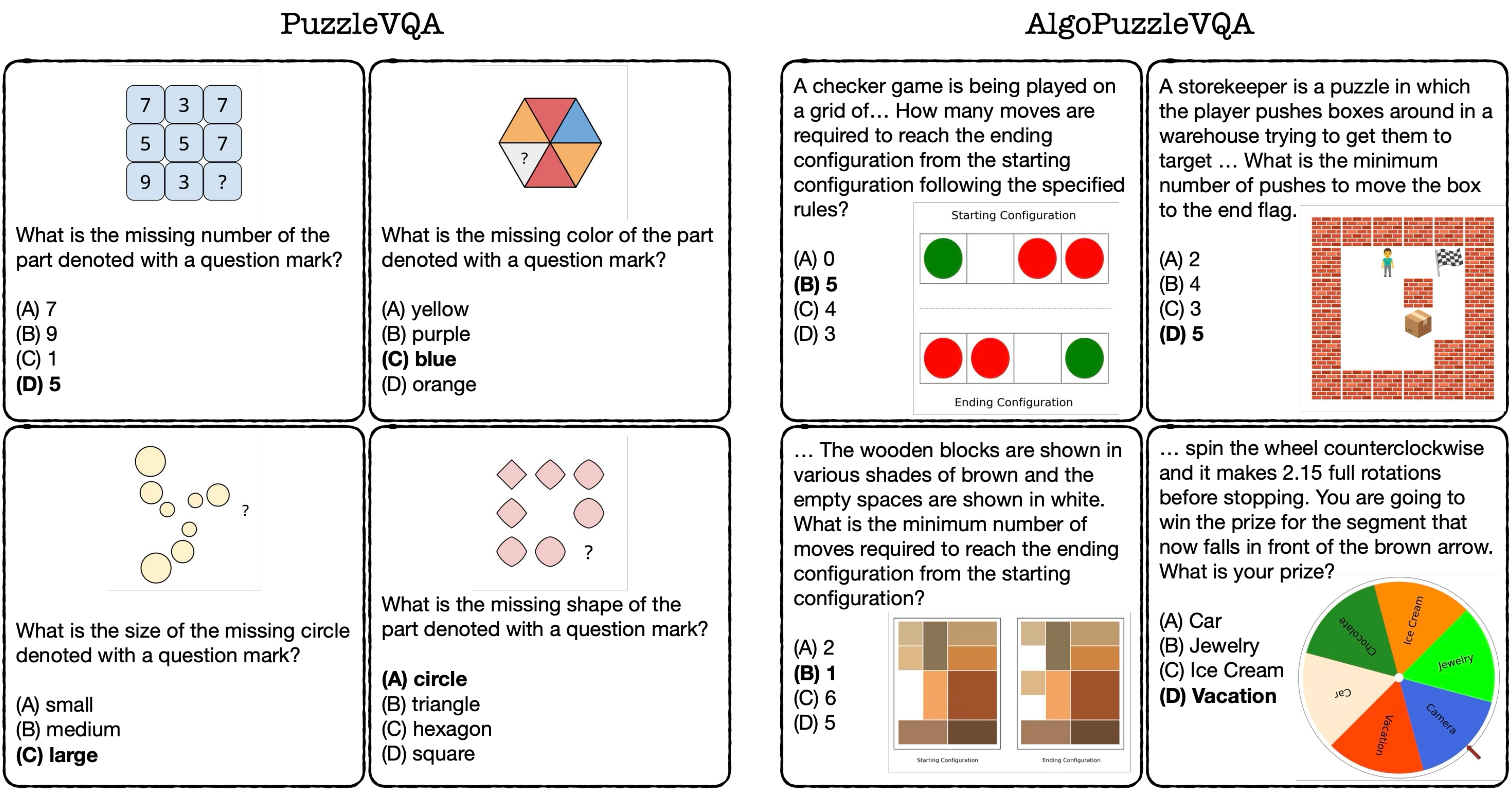}
    \caption{Examples of puzzles from \abstractdataset{} and \algodataset{}.}
    \label{fig:merged_examples}
\end{figure*}

\section{\abstractdataset{} \& \algodataset{}}

Understanding the capabilities and limitations of large multimodal models in reasoning tasks requires datasets that challenge their cognitive capabilities in nuanced ways. 
Multimodal puzzles are essential benchmarks for evaluating these models, as they require a unique combination of perception, reasoning, and abstraction.
In this study, we employ \abstractdataset{} which emphasizes visual abstract reasoning and pattern recognition, alongside \algodataset{}, which features more complex puzzles that require algorithmic solutions.

\abstractdataset{} consists 10 puzzle categories, with a total of 2,000 test instances. Four of these categories focus on single-concept patterns, such as numbers, colors, sizes, and shapes, while the remaining six categories emphasize dual-concept patterns, which combine two distinct concepts.
\algodataset{} consists of 18 distinct puzzles, each with 100 test instances, resulting in a
total of 1,800 test instances. These puzzles cover a wide range of topics, incorporating both visual elements like colors and positions, as well as algorithmic concepts such as Boolean logic and optimization.
\Cref{fig:merged_examples} presents example puzzles from both datasets.

A key contribution of our study is expanding both datasets beyond their original multiple-choice format to include open-ended format. This enhancement enables a more comprehensive and in-depth evaluation of model reasoning capabilities.
Additionally, we assess GPT-[n] and o-[n] models, which have not been examined in previous research. By doing so, we extend prior evaluations with a significantly deeper analysis, richer discussions, and extensive case study examples.

\section{Experimental Setup}

\subsection{Evaluation Pipeline}
For both multiple-choice and open-ended setups, we use GPT-4o for answer matching, as the varied output formats makes answer extraction a non-trivial task. The prompt used in this matching process is provided in \Cref{app:gpt4o_matching}. In both settings, we evaluate performance based on the accuracy of predicting the correct final answer.
To validate the reliability of GPT-4o for answer matching, we manually evaluated 200 random outputs with their matching scores. The results showed a classification accuracy of 99\%, with the task proving straightforward due to the simplicity of the final answers.

\subsection{Models}

We investigate the performance of GPT-[n] and o-[n] models: (1) GPT-4-Turbo, (2) GPT-4o, (3) GPT-4.5, (4) GPT-4.1, (5) o1, (6) o3, (7) o4-mini. 
We selected these two model series from OpenAI due to their rapid advancements and significant contributions to the field of LLMs. 
Each version has introduced innovative techniques that have shaped the LLM landscape. 
For example, GPT-4-Turbo excels in understanding visual inputs, while GPT-4o is optimized for efficient multimodal inference. 
The o-[n] models, a recent addition, is trained using reinforcement learning to perform complex reasoning before responding.
For the o-[n] models, we conducted evaluations across all three reasoning modes: \textit{``Low''}, \textit{``Medium''}, \textit{``High''}.
Please note that our study can easily be expanded to other closed-sourced and open-sourced models.

\begin{table*}[ht!]
\centering
\small
\setlength{\tabcolsep}{9pt}
\resizebox{\textwidth}{!}{%
\begin{tabular}{l|l|ccccccccccccc}
\toprule
& \multirow{2}{*}{\textbf{Tasks}} & \multirow{2}{*}{GPT-4-Turbo} & \multirow{2}{*}{GPT-4o} & \multirow{2}{*}{GPT-4.5} & \multirow{2}{*}{GPT-4.1} & \multicolumn{3}{c}{o1} & \multicolumn{3}{c}{o3} & \multicolumn{3}{c}{o4-mini} \\
\cmidrule(lr){7-9}\cmidrule(lr){10-12}\cmidrule(lr){13-15} 
& &  &  &  &  & Low & Medium & High & Low & Medium & High & Low & Medium & High \\
\midrule 
\multirow{32}{*}{\rotatebox[origin=c]{90}{ \textsc{\textbf{Multiple-Choice}} }} & \multicolumn{14}{c}{\cellcolor{gray!20}\textsc{PuzzleVQA}} \\
\greyrule{2}{15}
& Colors & 43.0 & 75.0 & 88.5 & 86.0 & 88.5 & 91.5 & 91.5 & 98.5 & 98.5 & 99.0 & 99.5 & 100.0 & 99.5 \\
& Numbers & 83.0 & 85.0 & 97.5 & 98.5 & 96.5 & 98.0 & 99.0 & 99.5 & 99.0 & 99.5 & 99.0 & 98.5 & 99.0 \\
& Shapes & 58.5 & 67.5 & 74.5 & 75.0 & 59.5 & 66.5 & 66.5 & 78.5 & 78.0 & 74.0 & 61.0 & 74.5 & 70.5 \\
& Size & 37.0 & 44.0 & 70.5 & 61.0 & 63.5 & 70.5 & 77.5 & 81.0 & 85.5 & 87.5 & 79.0 & 78.5 & 86.0 \\
\greyrule{2}{15}
& Colors \& Numbers & 62.0 & 52.0 & 86.5 & 82.5 & 92.5 & 98.5 & 99.0 & 95.0 & 97.5 & 97.5 & 99.0 & 100.0 & 99.5 \\
& Colors \& Shapes & 61.5 & 64.0 & 78.5 & 64.0 & 69.5 & 76.5 & 80.0 & 87.5 & 95.5 & 92.0 & 79.5 & 86.5 & 86.5 \\
& Colors \& Size & 48.0 & 57.5 & 60.5 & 65.5 & 43.0 & 49.0 & 50.0 & 53.5 & 58.5 & 57.0 & 64.5 & 67.5 & 72.5 \\
& Numbers \& Shapes & 51.5 & 42.5 & 43.0 & 67.0 & 82.0 & 86.5 & 92.0 & 88.0 & 88.0 & 92.0 & 87.5 & 93.0 & 96.0 \\
& Numbers \& Size & 30.5 & 29.5 & 47.5 & 52.0 & 45.0 & 44.0 & 47.0 & 48.5 & 49.0 & 45.0 & 42.0 & 64.0 & 62.0 \\
& Size \& Shapes & 55.0 & 60.5 & 95.0 & 70.0 & 72.0 & 82.5 & 86.5 & 91.0 & 91.5 & 97.0 & 78.5 & 86.5 & 85.5 \\
\greyrule{2}{15} 
& \cellcolor{blue!15}\textbf{Average} & \cellcolor{blue!15}53.0 & \cellcolor{blue!15}57.8 & \cellcolor{blue!15}74.2 & \cellcolor{blue!15}72.2 & \cellcolor{blue!15}71.2 & \cellcolor{blue!15}76.3 & \cellcolor{blue!15}78.9 & \cellcolor{blue!15}82.1 & \cellcolor{blue!15}84.1 & \cellcolor{blue!15}84.0 & \cellcolor{blue!15}79.0 & \cellcolor{blue!15}84.9 & \cellcolor{blue!15}85.7 \\

\greyrule{2}{15}
 & \multicolumn{14}{c}{\cellcolor{gray!20}\textsc{AlgoPuzzleVQA}} \\
\greyrule{2}{15}
& Board Tiling & 49.0 & 52.0 & 51.0 & 55.0 & 52.0 & 53.0 & 48.0 & 55.0 & 60.0 & 60.0 & 48.0 & 44.0 & 52.0 \\
& Calendar & 55.0 & 61.0 & 67.0 & 68.0 & 84.0 & 90.0 & 90.0 & 93.0 & 96.0 & 96.0 & 94.0 & 98.0 & 99.0 \\
& Chain Link & 4.0 & 5.0 & 2.0 & 2.0 & 30.0 & 35.0 & 58.0 & 47.0 & 65.0 & 75.0 & 33.0 & 45.0 & 77.0 \\
& Checker Move & 15.0 & 18.0 & 36.0 & 34.0 & 41.0 & 46.0 & 52.0 & 61.0 & 59.0 & 56.0 & 82.0 & 94.0 & 97.0 \\
& Clock & 19.0 & 14.0 & 56.0 & 68.0 & 65.0 & 76.0 & 81.0 & 96.0 & 97.0 & 98.0 & 85.0 & 92.0 & 91.0 \\
& Colour Hue & 33.0 & 25.0 & 23.0 & 26.0 & 36.0 & 43.0 & 43.0 & 41.0 & 36.0 & 47.0 & 42.0 & 44.0 & 47.0 \\
& Map Colour & 20.0 & 28.0 & 0.0 & 0.0 & 3.0 & 1.0 & 50.0 & 2.0 & 4.0 & 5.0 & 3.0 & 4.0 & 3.0 \\
& Maze Solve & 32.0 & 39.0 & 30.0 & 27.0 & 41.0 & 44.0 & 50.0 & 36.0 & 39.0 & 39.0 & 39.0 & 32.0 & 42.0 \\
& Move Box & 34.0 & 35.0 & 41.0 & 37.0 & 26.0 & 24.0 & 30.0 & 49.0 & 46.0 & 52.0 & 46.0 & 47.0 & 50.0 \\
& N-Queens & 18.0 & 14.0 & 3.0 & 3.0 & 18.0 & 13.0 & 15.0 & 34.0 & 38.0 & 35.0 & 39.0 & 52.0 & 56.0 \\
& Number Slide & 27.0 & 32.0 & 21.0 & 18.0 & 26.0 & 27.0 & 88.0 & 28.0 & 27.0 & 30.0 & 29.0 & 29.0 & 29.0 \\
& Rotten Fruits & 25.0 & 53.0 & 81.0 & 72.0 & 43.0 & 48.0 & 52.0 & 86.0 & 84.0 & 84.0 & 96.0 & 99.0 & 97.0 \\
& Rubik's Cube & 40.0 & 31.0 & 61.0 & 47.0 & 64.0 & 68.0 & 74.0 & 78.0 & 75.0 & 85.0 & 77.0 & 80.0 & 92.0 \\
& Think A Dot & 37.0 & 42.0 & 40.0 & 49.0 & 53.0 & 57.0 & 61.0 & 66.0 & 74.0 & 69.0 & 67.0 & 82.0 & 74.0 \\
& Tower of Hanoi & 8.0 & 19.0 & 29.0 & 39.0 & 50.0 & 59.0 & 63.0 & 77.0 & 81.0 & 79.0 & 60.0 & 76.0 & 85.0 \\
& Water Jugs & 13.0 & 34.0 & 37.0 & 43.0 & 48.0 & 57.0 & 43.0 & 80.0 & 89.0 & 88.0 & 67.0 & 82.0 & 82.0 \\
& Wheel of Fortune & 15.0 & 33.0 & 47.0 & 51.0 & 48.0 & 55.0 & 56.0 & 66.0 & 65.0 & 75.0 & 50.0 & 58.0 & 69.0 \\
& Wood Slide & 13.0 & 21.0 & 22.0 & 28.0 & 26.0 & 23.0 & 25.0 & 46.0 & 52.0 & 48.0 & 33.0 & 24.0 & 36.0 \\
\greyrule{2}{15}
& \cellcolor{blue!15}\textbf{Average} & \cellcolor{blue!15}25.4 & \cellcolor{blue!15}30.9 & \cellcolor{blue!15}35.9 & \cellcolor{blue!15}37.1 & \cellcolor{blue!15}41.9 & \cellcolor{blue!15}45.5 & \cellcolor{blue!15}54.4 & \cellcolor{blue!15}57.8 & \cellcolor{blue!15}60.4 & \cellcolor{blue!15}62.3 & \cellcolor{blue!15}55.0 & \cellcolor{blue!15}60.1 & \cellcolor{blue!15}65.4 \\
\bottomrule
\toprule

\multirow{32}{*}{\rotatebox[origin=c]{90}{ \textsc{\textbf{Open-Ended}} }} & \multicolumn{14}{c}{\cellcolor{gray!20}\textsc{PuzzleVQA}} \\
\greyrule{2}{15}
& Colors & 51.0 & 72.5 & 81.0 & 87.0 & 89.5 & 90.0 & 80.5 & 98.5 & 98.5 & 98.0 & 99.5 & 100.0 & 99.5 \\
& Numbers & 82.5 & 84.5 & 91.0 & 95.5 & 93.0 & 98.0 & 96.5 & 95.0 & 94.5 & 93.0 & 93.0 & 91.5 & 93.0 \\
& Shapes & 32.5 & 51.5 & 66.5 & 54.0 & 45.0 & 48.0 & 54.5 & 63.0 & 62.0 & 62.5 & 57.0 & 55.0 & 57.0 \\
& Size & 19.0 & 39.0 & 53.0 & 59.5 & 48.5 & 55.0 & 54.5 & 72.0 & 75.0 & 72.0 & 57.5 & 69.0 & 79.5 \\
\greyrule{2}{15}
& Colors \& Numbers & 54.5 & 48.0 & 71.0 & 67.5 & 93.0 & 96.5 & 97.0 & 90.5 & 93.5 & 92.0 & 97.0 & 97.5 & 98.0 \\
& Colors \& Shapes & 30.0 & 45.5 & 75.5 & 48.5 & 67.5 & 74.0 & 75.0 & 89.0 & 90.5 & 90.0 & 74.0 & 84.0 & 88.5 \\
& Colors \& Size & 31.5 & 21.5 & 55.0 & 54.5 & 35.5 & 35.5 & 30.0 & 48.0 & 55.5 & 57.0 & 49.5 & 57.5 & 59.0 \\
& Numbers \& Shapes & 31.5 & 20.0 & 26.0 & 40.0 & 67.0 & 78.5 & 78.0 & 73.5 & 78.5 & 83.5 & 66.5 & 71.5 & 86.0 \\
& Numbers \& Size & 24.5 & 34.5 & 43.0 & 42.0 & 38.5 & 44.5 & 41.5 & 43.5 & 43.5 & 47.0 & 38.0 & 45.0 & 52.0 \\
& Size \& Shapes & 28.5 & 50.5 & 74.0 & 56.0 & 47.0 & 53.0 & 55.0 & 77.5 & 78.0 & 81.0 & 76.5 & 78.0 & 91.0 \\
\greyrule{2}{15}
& \cellcolor{blue!15}\textbf{Average} & \cellcolor{blue!15}38.5 & \cellcolor{blue!15}46.8 & \cellcolor{blue!15}63.6 & \cellcolor{blue!15}60.5 & \cellcolor{blue!15}62.5 & \cellcolor{blue!15}67.3 & \cellcolor{blue!15}66.2 & \cellcolor{blue!15}75.0 & \cellcolor{blue!15}77.0 & \cellcolor{blue!15}77.6 & \cellcolor{blue!15}70.8 & \cellcolor{blue!15}74.9 & \cellcolor{blue!15}80.3 \\

\greyrule{2}{15}
 & \multicolumn{14}{c}{\cellcolor{gray!20}\textsc{AlgoPuzzleVQA}} \\
\greyrule{2}{15}
& Board Tiling & 46.0 & 46.0 & 56.0 & 51.0 & 52.0 & 49.0 & 51.0 & 58.0 & 58.0 & 61.0 & 48.0 & 51.0 & 53.0 \\
& Calendar & 43.0 & 52.0 & 64.0 & 63.0 & 71.0 & 80.0 & 83.0 & 85.0 & 95.0 & 94.0 & 90.0 & 93.0 & 97.0 \\
& Chain Link & 1.0 & 3.0 & 0.0 & 0.0 & 0.0 & 0.0 & 1.0 & 3.0 & 2.0 & 4.0 & 0.0 & 1.0 & 6.0 \\
& Checker Move & 3.0 & 7.0 & 14.0 & 18.0 & 28.0 & 28.0 & 34.0 & 46.0 & 49.0 & 46.0 & 32.0 & 52.0 & 74.0 \\
& Clock & 0.0 & 3.0 & 12.0 & 20.0 & 6.0 & 6.0 & 6.0 & 19.0 & 16.0 & 25.0 & 20.0 & 15.0 & 13.0 \\
& Colour Hue & 5.0 & 10.0 & 15.0 & 14.0 & 14.0 & 15.0 & 15.0 & 20.0 & 17.0 & 23.0 & 12.0 & 18.0 & 20.0 \\
& Map Colour & 10.0 & 22.0 & 0.0 & 1.0 & 1.0 & 0.0 & 21.0 & 1.0 & 0.0 & 1.0 & 2.0 & 2.0 & 1.0 \\
& Maze Solve & 16.0 & 8.0 & 9.0 & 14.0 & 14.0 & 22.0 & 17.0 & 14.0 & 11.0 & 10.0 & 15.0 & 15.0 & 16.0 \\
& Move Box & 20.0 & 23.0 & 32.0 & 43.0 & 24.0 & 25.0 & 23.0 & 40.0 & 47.0 & 47.0 & 34.0 & 39.0 & 41.0 \\
& N-Queens & 17.0 & 16.0 & 12.0 & 8.0 & 12.0 & 12.0 & 16.0 & 21.0 & 23.0 & 22.0 & 30.0 & 31.0 & 47.0 \\
& Number Slide & 14.0 & 32.0 & 19.0 & 17.0 & 26.0 & 23.0 & 71.0 & 27.0 & 27.0 & 28.0 & 23.0 & 27.0 & 28.0 \\
& Rotten Fruits & 32.0 & 53.0 & 69.0 & 71.0 & 41.0 & 41.0 & 43.0 & 75.0 & 76.0 & 77.0 & 95.0 & 95.0 & 96.0 \\
& Rubik's Cube & 32.0 & 44.0 & 58.0 & 48.0 & 52.0 & 53.0 & 54.0 & 68.0 & 63.0 & 69.0 & 66.0 & 72.0 & 78.0 \\
& Think A Dot & 36.0 & 41.0 & 41.0 & 44.0 & 40.0 & 45.0 & 32.0 & 38.0 & 48.0 & 38.0 & 55.0 & 56.0 & 55.0 \\
& Tower of Hanoi & 0.0 & 2.0 & 10.0 & 24.0 & 31.0 & 35.0 & 39.0 & 60.0 & 69.0 & 60.0 & 38.0 & 60.0 & 64.0 \\
& Water Jugs & 8.0 & 23.0 & 24.0 & 34.0 & 29.0 & 43.0 & 42.0 & 72.0 & 85.0 & 87.0 & 53.0 & 66.0 & 76.0 \\
& Wheel of Fortune & 14.0 & 29.0 & 33.0 & 36.0 & 37.0 & 34.0 & 31.0 & 43.0 & 43.0 & 49.0 & 33.0 & 34.0 & 38.0 \\
& Wood Slide & 0.0 & 1.0 & 1.0 & 9.0 & 0.0 & 0.0 & 0.0 & 16.0 & 25.0 & 29.0 & 0.0 & 4.0 & 9.0 \\
\greyrule{2}{15}
& \cellcolor{blue!15}\textbf{Average} & \cellcolor{blue!15}16.5 & \cellcolor{blue!15}23.1 & \cellcolor{blue!15}26.1 & \cellcolor{blue!15}28.6 & \cellcolor{blue!15}26.6 & \cellcolor{blue!15}28.4 & \cellcolor{blue!15}32.2 & \cellcolor{blue!15}39.2 & \cellcolor{blue!15}41.9 & \cellcolor{blue!15}42.8 & \cellcolor{blue!15}35.9 & \cellcolor{blue!15}40.6 & \cellcolor{blue!15}45.1 \\

\bottomrule
\end{tabular}
}
\caption{Performance comparison of GPT-[n] and o-[n] models on the \textsc{PuzzleVQA} and \textsc{AlgoPuzzleVQA} datasets under both multiple-choice and open-ended settings.}
\label{tab:results}
\end{table*}

\section{Overall Performance Trends and Model Comparisons}
The results presented in \Cref{tab:results} provides a comprehensive comparison of the current capabilities and limitations of GPT-[n] and o-[n] models across \textsc{PuzzleVQA} and \textsc{AlgoPuzzleVQA}. Evaluations conducted under both multiple-choice and open-ended settings reveal clear performance tiers, showcasing notable advancements while also highlighting persistent challenges.

\begin{figure*}[ht]
    \centering
    \includegraphics[width=1.0\linewidth]{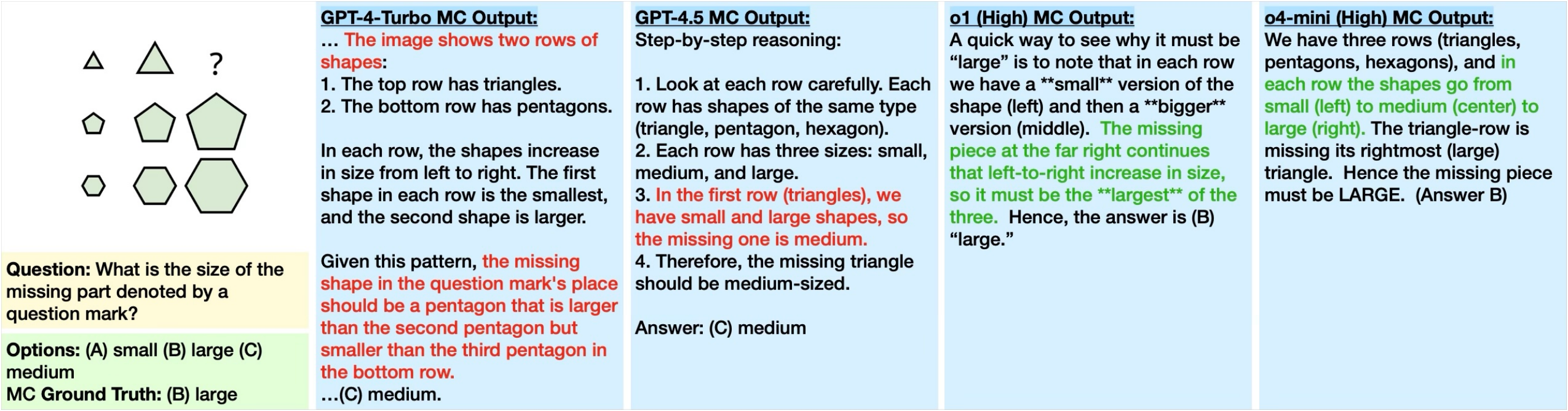}
    \caption{A case study on the \textit{Size \& Shape} puzzle in \abstractdataset{} reveals that GPT-[n] models exhibit weaker visual perception capabilities compared to o-[n] models.}
    \label{fig:case_shape_size}
\end{figure*}

\paragraph{Superior Performance and Scalability of o-[n] Models.}
Across both settings, o-[n] models consistently outperform GPT-[n] models by a huge margin on both datasets. Within the o-[n] series, performance scales reliably with increased reasoning modes: \textit{``High''} variants outperform \textit{``Medium''}, which in turn surpass \textit{``Low''}. This trend highlights the scalability of the o-[n] architecture and training strategy, effectively converting additional resources into better performance. Furthermore, newer generations within the series, notably o3 and especially o4-mini, set new benchmarks, underscoring continued progress.

\paragraph{GPT-[n] Models Progression and Limitations.}
Among the GPT-[n] series, GPT-4.5 and GPT-4.1 generally outperform earlier versions like GPT-4o, reflecting steady improvements across iterations. However, even these newer models are consistently outperformed by o-[n] models across a broad range of tasks. This persistent gap highlights the effectiveness of the o-[n] series' training strategies, particularly for tasks demanding complex visual understanding and abstract or algorithmic reasoning. A case study in \Cref{fig:case_shape_size} illustrates this disparity, showing how current GPT-[n] models fall short in visual perception compared to o-[n] models.

\begin{figure*}
    \centering
    \includegraphics[width=1\linewidth]{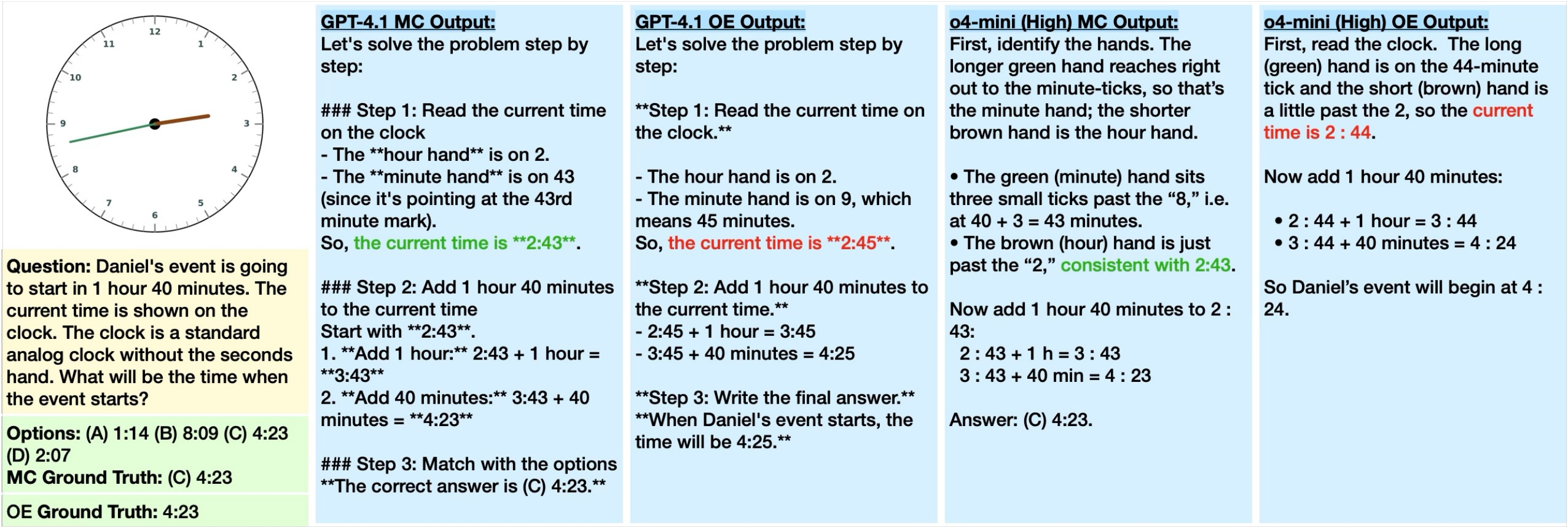}
    \caption{A case study on the \textit{Clock} puzzle in \algodataset{}, evaluated in both multiple-choice and open-ended settings, shows that both GPT-[n] and o-[n] models still lack fine-grained visual perception capabilities.}
    \label{fig:case_clock}
\end{figure*}

\paragraph{Impact of Evaluation Settings and Dataset Complexity.}
The format of evaluation—multiple-choice versus open-ended—has a significant impact on model performance. Models consistently perform better on multiple-choice tasks, where the limited answer space enables success through recognition or elimination rather than generative recall. In contrast, open-ended questions, which require precise, unaided responses, result in a drop in accuracy. This gap underscores persistent challenges in achieving unconstrained, fine-grained reasoning. A case study in \Cref{fig:case_clock} illustrates this: models often fail to accurately infer the time from a clock image, reflecting difficulties with detailed visual perception. Dataset complexity further amplifies these differences. The \algodataset{} dataset, which emphasizes algorithmic, spatial, and multi-step reasoning, consistently proves more difficult than the \textsc{PuzzleVQA} dataset, which focuses on visual attributes and simpler logic. All models perform worse on \textsc{AlgoPuzzleVQA}, with the performance gap between o-[n] and GPT-[n] models widening under increased complexity.

\section{Performance on \textsc{PuzzleVQA}}

\paragraph{Strong Performance on Basic Concepts}
On \textsc{PuzzleVQA}, particularly for tasks centered around singular concepts like \textit{Colors} or \textit{Numbers}, the more advanced o-[n] models, demonstrate exceptional proficiency, frequently achieving near-perfect or perfect scores (e.g., 98-100\% for o4-mini (High) on \textit{Colors} and \textit{Numbers} in multiple-choice). This indicates a very strong grasp of fundamental visual perception and attribute extraction. GPT-[n] models, particularly GPT-4.5 and GPT-4.1, also perform reasonably well on these simpler tasks, though generally not reaching the same consistent near-perfection as o-[n] models.

\paragraph{Challenges with Concept Combinations and Size Perception}
The challenge becomes more pronounced when tasks require reasoning over combinations of concepts such as \textit{Colors \& Shapes} or \textit{Numbers \& Size} or involve perceiving \textit{Size} alone. In these puzzles, all models exhibit a noticeable decline in performance, underscoring the inherent difficulty of compositional reasoning, where multiple visual elements must be integrated accurately. Among these concepts, \textit{Size} emerges as a consistent bottleneck, whether in isolation, or in conjunction with others (\textit{Colors \& Size}, \textit{Numbers \& Size}) tend to achieve lower scores than those focused solely on \textit{Colors} or \textit{Numbers}. This suggests that current models struggle more with interpreting relative spatial extent and making proportional comparisons than with processing discrete, categorical features. Nevertheless, the o-[n] models continues to outperform others, demonstrating stronger capabilities in compositional and relational visual reasoning.

\paragraph{Average Performance Summary for \textsc{PuzzleVQA}}
In the multiple-choice setting, GPT-4.5 achieved an average accuracy of 74.2\%. In comparison, the o4-mini (High) model performed significantly better, with an average of 85.7\%. Other high-performing models in the o-[n] family, such as o3 (Medium/High), also scored in the mid-80s. A similar performance gap is evident in the open-ended setting, GPT-4.5 averaged 63.6\%, whereas o4-mini (High) reached an impressive 80.3\%. These results underscore the o-[n] models’ superior ability to comprehend and reason through the visual puzzles in the dataset, demonstrating strength not only in basic visual perception but also in understanding more complex concept combinations. Several case study examples are presented in \Cref{app:abstract_case_study}.

\begin{table*}[!ht]
\centering
\small
\setlength{\tabcolsep}{9pt}
\resizebox{\textwidth}{!}{%
\begin{tabular}{l|l|ccccccccccccc}
\toprule
& \multirow{2}{*}{\textbf{Tasks}} & \multirow{2}{*}{GPT-4-Turbo} & \multirow{2}{*}{GPT-4o} & \multirow{2}{*}{GPT-4.5} & \multirow{2}{*}{GPT-4.1} & \multicolumn{3}{c}{o1} & \multicolumn{3}{c}{o3} & \multicolumn{3}{c}{o4-mini} \\
\cmidrule(lr){7-9}\cmidrule(lr){10-12}\cmidrule(lr){13-15}
& &  &  &  &  & Low & Medium & High & Low & Medium & High & Low & Medium & High \\
\midrule 
\multirow{36}{*}{\rotatebox[origin=c]{90}{ \textsc{\textbf{Open-Ended PuzzleVQA}} }} & \multicolumn{14}{c}{\cellcolor{gray!20}\textsc{Original}} \\
\greyrule{2}{15}
& Colors & 51.0 & 72.5 & 81.0 & 87.0 & 89.5 & 90.0 & 80.5 & 98.5 & 98.5 & 98.0 & 99.5 & 100.0 & 99.5 \\
& Numbers & 82.5 & 84.5 & 91.0 & 95.5 & 93.0 & 98.0 & 96.5 & 95.0 & 94.5 & 93.0 & 93.0 & 91.5 & 93.0 \\
& Shapes & 32.5 & 51.5 & 66.5 & 54.0 & 45.0 & 48.0 & 54.5 & 63.0 & 62.0 & 62.5 & 57.0 & 55.0 & 57.0 \\
& Size & 19.0 & 39.0 & 53.0 & 59.5 & 48.5 & 55.0 & 54.5 & 72.0 & 75.0 & 72.0 & 57.5 & 69.0 & 79.5 \\
\greyrule{2}{15}
& Colors \& Numbers & 54.5 & 48.0 & 71.0 & 67.5 & 93.0 & 96.5 & 97.0 & 90.5 & 93.5 & 92.0 & 97.0 & 97.5 & 98.0 \\
& Colors \& Shapes & 30.0 & 45.5 & 75.5 & 48.5 & 67.5 & 74.0 & 75.0 & 89.0 & 90.5 & 90.0 & 74.0 & 84.0 & 88.5 \\
& Colors \& Size & 31.5 & 21.5 & 55.0 & 54.5 & 35.5 & 35.5 & 30.0 & 48.0 & 55.5 & 57.0 & 49.5 & 57.5 & 59.0 \\
& Numbers \& Shapes & 31.5 & 20.0 & 26.0 & 40.0 & 67.0 & 78.5 & 78.0 & 73.5 & 78.5 & 83.5 & 66.5 & 71.5 & 86.0 \\
& Numbers \& Size & 24.5 & 34.5 & 43.0 & 42.0 & 38.5 & 44.5 & 41.5 & 43.5 & 43.5 & 47.0 & 38.0 & 45.0 & 52.0 \\
& Size \& Shapes & 28.5 & 50.5 & 74.0 & 56.0 & 47.0 & 53.0 & 55.0 & 77.5 & 78.0 & 81.0 & 76.5 & 78.0 & 91.0 \\
\greyrule{2}{15}
& \cellcolor{blue!15}\textbf{Average} & \cellcolor{blue!15}38.5 & \cellcolor{blue!15}46.8 & \cellcolor{blue!15}63.6 & \cellcolor{blue!15}60.5 & \cellcolor{blue!15}62.5 & \cellcolor{blue!15}67.3 & \cellcolor{blue!15}66.2 & \cellcolor{blue!15}75.0 & \cellcolor{blue!15}77.0 & \cellcolor{blue!15}77.6 & \cellcolor{blue!15}70.8 & \cellcolor{blue!15}74.9 & \cellcolor{blue!15}80.3 \\
\greyrule{2}{15}

& \multicolumn{14}{c}{\cellcolor{gray!20}\textsc{Original + Visual Perception}} \\
\greyrule{2}{15}
& Colors & 75.0 & 80.0 & 92.0 & 95.5 & 99.0 & 100.0 & 94.0 & 100.0 & 100.0 & 100.0 & 100.0 & 100.0 & 100.0 \\
& Numbers & 77.0 & 88.5 & 98.0 & 94.5 & 98.5 & 99.5 & 98.0 & 95.5 & 92.5 & 91.5 & 97.5 & 92.5 & 92.0 \\
& Shapes & 71.5 & 63.5 & 66.0 & 63.5 & 45.5 & 50.5 & 55.5 & 68.0 & 63.5 & 65.5 & 54.5 & 54.5 & 54.0 \\
& Size & 64.5 & 62.5 & 93.5 & 97.0 & 93.5 & 98.0 & 98.0 & 98.5 & 99.5 & 99.0 & 99.0 & 99.5 & 100.0 \\
\greyrule{2}{15}
& Colors \& Numbers & 67.0 & 52.0 & 94.5 & 83.5 & 97.0 & 95.0 & 95.0 & 95.5 & 96.5 & 94.5 & 96.5 & 98.0 & 98.5 \\
& Colors \& Shapes & 81.0 & 77.5 & 75.5 & 70.0 & 86.5 & 82.5 & 81.5 & 86.5 & 88.5 & 86.5 & 82.5 & 83.0 & 89.5 \\
& Colors \& Size & 53.5 & 78.0 & 84.5 & 80.5 & 96.0 & 99.0 & 99.0 & 96.5 & 99.0 & 97.0 & 93.0 & 98.0 & 99.5 \\
& Numbers \& Shapes & 29.5 & 33.5 & 26.5 & 35.0 & 88.5 & 89.0 & 86.0 & 78.0 & 82.0 & 83.0 & 79.5 & 86.5 & 90.5 \\
& Numbers \& Size & 70.0 & 73.0 & 79.0 & 68.5 & 77.0 & 82.0 & 81.5 & 81.0 & 82.5 & 79.5 & 69.0 & 80.0 & 77.5 \\
& Size \& Shapes & 97.5 & 92.5 & 92.5 & 88.0 & 92.0 & 98.0 & 98.0 & 93.5 & 94.5 & 96.5 & 90.0 & 98.5 & 98.5 \\
\greyrule{2}{15}
& \cellcolor{blue!15}\textbf{Average} & \cellcolor{blue!15}68.7 & \cellcolor{blue!15}70.1 & \cellcolor{blue!15}80.2 & \cellcolor{blue!15}77.6 & \cellcolor{blue!15}87.3 & \cellcolor{blue!15}89.3 & \cellcolor{blue!15}88.7 & \cellcolor{blue!15}89.3 & \cellcolor{blue!15}89.8 & \cellcolor{blue!15}89.3 & \cellcolor{blue!15}86.2 & \cellcolor{blue!15}89.0 & \cellcolor{blue!15}90.0 \\
\greyrule{2}{15}

& \multicolumn{14}{c}{\cellcolor{gray!20}\textsc{Original + Visual Perception + Induction}} \\
\greyrule{2}{15}
& Colors & 97.0 & 92.0 & 100.0 & 100.0 & 100.0 & 100.0 & 99.0 & 99.5 & 100.0 & 100.0 & 100.0 & 100.0 & 100.0 \\
& Numbers & 98.5 & 99.5 & 98.5 & 100.0 & 98.0 & 96.5 & 97.0 & 100.0 & 100.0 & 99.5 & 99.5 & 99.0 & 98.5 \\
& Shapes & 97.5 & 97.5 & 100.0 & 93.0 & 97.5 & 99.0 & 100.0 & 88.5 & 88.0 & 91.0 & 91.5 & 91.0 & 94.0 \\
& Size & 95.5 & 96.5 & 99.0 & 100.0 & 100.0 & 100.0 & 100.0 & 100.0 & 100.0 & 100.0 & 100.0 & 99.5 & 100.0 \\
\greyrule{2}{15}
& Colors \& Numbers & 89.5 & 89.5 & 98.5 & 99.5 & 100.0 & 100.0 & 100.0 & 99.5 & 100.0 & 100.0 & 99.5 & 99.0 & 99.5 \\
& Colors \& Shapes & 64.5 & 77.0 & 88.5 & 87.5 & 88.0 & 87.0 & 89.5 & 95.5 & 93.5 & 97.5 & 90.5 & 94.5 & 98.0 \\
& Colors \& Size & 75.5 & 94.5 & 93.5 & 86.5 & 92.5 & 92.5 & 94.0 & 98.0 & 98.0 & 99.0 & 99.0 & 98.0 & 99.0 \\
& Numbers \& Shapes & 84.5 & 85.5 & 78.5 & 84.0 & 91.0 & 89.5 & 91.0 & 90.0 & 88.5 & 87.5 & 93.0 & 93.0 & 93.5 \\
& Numbers \& Size & 63.0 & 73.5 & 78.5 & 87.0 & 74.0 & 72.5 & 77.5 & 80.0 & 82.0 & 81.5 & 63.5 & 57.0 & 60.5 \\
& Size \& Shapes & 93.0 & 92.5 & 90.0 & 91.5 & 97.5 & 99.0 & 99.5 & 95.5 & 97.5 & 95.0 & 98.0 & 98.5 & 99.5 \\
\greyrule{2}{15}
& \cellcolor{blue!15}\textbf{Average} & \cellcolor{blue!15}85.8 & \cellcolor{blue!15}89.8 & \cellcolor{blue!15}92.5 & \cellcolor{blue!15}92.9 & \cellcolor{blue!15}93.8 & \cellcolor{blue!15}93.6 & \cellcolor{blue!15}94.8 & \cellcolor{blue!15}94.7 & \cellcolor{blue!15}94.8 & \cellcolor{blue!15}95.1 & \cellcolor{blue!15}93.5 & \cellcolor{blue!15}93.0 & \cellcolor{blue!15}94.2 \\

\bottomrule
\end{tabular}
}
\caption{Bottleneck analysis of GPT-[n] and o-[n] models on \abstractdataset{} in the open-ended setting. \textbf{Original} refers to our default setting where only a question and an image are provided as input. To reveal the specific multimodal reasoning bottlenecks, we progressively inject ground-truth explanations in the input for visual perception and inductive reasoning. We provide an example of the different prompts used in the bottleneck analysis in \Cref{fig:bottleneck}.}
\label{tab:bottleneck}
\end{table*}

\section{Performance on \textsc{AlgoPuzzleVQA}}

\paragraph{Significant Challenges in Algorithmic and Spatial Reasoning}
The results on \textsc{AlgoPuzzleVQA} underscore that complex algorithmic and spatial reasoning remains a challenge for current models. Overall scores are considerably lower across all models compared to \textsc{PuzzleVQA}.
GPT-[n] models struggle significantly with certain tasks such as \textit{Chain Link} and \textit{Map Colour} in the open-ended setting, often scoring near zero, indicating a fundamental difficulty in undestanding the underlying principles or performing necessary reasoning steps for these puzzle types.

\paragraph{Gains by o-[n] Models on Complex Algorithmic Tasks}
Despite the overall difficulty, the o-[n] models demonstrate improvements on several complex algorithmic tasks within \textsc{AlgoPuzzleVQA}. For instance, on \textit{Calendar} related puzzles, o4-mini (High) achieves scores in the high 90s (MC: 99.0\%, OE: 97.0\%), a huge difference compared to GPT-4.5's 67.0\% (MC) and 64.0\% (OE). 
Similarly, on \textit{Checker Move}, GPT-4.5 scores 36.0\% (MC), while o4-mini (High) reaches 97.0\% (MC). 
These improvements suggest that the o-[n] models have developed more advanced capabilities for sequential decision-making, state tracking, and applying learned algorithmic patterns, enabling them to tackle certain classes of complex problems with significantly greater effectiveness.

\paragraph{Challenges in Highly Abstract or Combinatorics Puzzles}
Puzzles such as \textit{Map Colour} and \textit{Chain Link} particularly in their open-ended formats consistently produce very low scores, with o4-mini High scoring only 1.0\% and GPT-4.5 0.0\% on \textit{Map Colour}, and 6.0\% and 0.0\% respectively on \textit{Chain Link}. These problems typically require abstract graph reasoning, intricate constraint satisfaction, or planning over a vast combinatorial space where even the most advanced models still struggle. 
Similarly, \textit{Wood Slide} remains a challenge in its open-ended format for all models.

\paragraph{Average Performance Summary for \textsc{AlgoPuzzleVQA}}
The average performance highlight both the dataset’s difficulty and the varying capabilities of different models. 
In the multiple-choice setting, the best-performing GPT model (GPT-4.1) achieved an average accuracy of 37.1\%, whereas o4-mini (High) significantly outperformed it with 65.4\%. A similar pattern holds for open-ended setting where GPT-4.1 reached 28.6\%, while o4-mini (High) attained 45.1\%. These results show that while the o-[n] models have made notable progress in tackling algorithmic puzzles, overall performance remains modest, indicating that this domain is far from being solved. We present several case study examples in \Cref{app:algo_case_study}.

\section{Bottleneck Analysis}
Our bottleneck analysis on \abstractdataset{} (\Cref{tab:bottleneck}) reveals that providing ground-truth visual perception shows performance gains across all models (GPT-4.5: 63.6\% to 80.2\%; o4-mini (High): 80.3\% to 90.0\%). This consistent improvement, observed across different models, clearly isolates visual perception as a primary bottleneck in the baseline. Subsequently, even with perfect visual perception assumed, injecting ground-truth inductive reasoning provides further significant performance increases (GPT-4.5: 80.2\% to 92.5\%; o4-mini (High): 90.0\% to 94.2\%), underscoring inductive reasoning as a separate and significant limiting factor. Overall, the progressive injection from the "Original" setting to "Original + Visual Perception + Induction" quantifies the substantial impact of these distinct components (GPT-4.5 total gain from 63.6\% to 92.5\%), indicating that while models possess strong underlying reasoning, their performance is notably constrained by difficulties in both visual perception and inductive reasoning. We present several case studies in \Cref{app:bottleneck_case_study}.

\section{Conclusion}
This study assesses the multimodal reasoning capabilities of the GPT-[n] and o-[n] model series using the \textsc{PuzzleVQA} and \textsc{AlgoPuzzleVQA} benchmarks, revealing a significant advancement in performance, particularly with later o-[n] iterations, which consistently and scalably outperform their GPT-[n] models. Across all models, performance was notably stronger in multiple-choice settings compared to open-ended ones. The o-[n] models demonstrated distinct strengths in specific algorithmic tasks (e.g., \textit{Calendar}) and basic concept recognition (e.g., \textit{Colors}). However, persistent challenges remain, particularly in compositional reasoning (e.g., \textit{Colors \& Shapes}), fine-grained visual perception (especially \textit{Size}), and complex algorithmic or combinatorial puzzles. Our bottleneck analysis highlights visual perception and subsequent inductive reasoning as the primary limiting factors—emphasizing that, despite architectural progress, substantial improvements in these core areas are essential to closing the gap toward robust, human-like multimodal intelligence.

\section{Limitations}
In this study, we conducted a thorough analysis of the performance of GPT-[n] and o-[n] models, offering detailed discussions and extensive case study examples to highlight their current challenges.
However, despite our best efforts, some limitations may still be present in this paper.
Due to space constraints, we were unable to analyze all existing multimodal puzzles comprehensively and instead focused primarily on \abstractdataset{} and \algodataset{}.
Additionally, our interpretations and conclusions are shaped by our own perspectives and understanding of the field. Other researchers may interpret the same studies differently.
Nevertheless, we believe this study provides valuable insights into the reasoning performance of GPT-[n] and o-[n] models on multimodal puzzles.

\section{Ethical Considerations}
It’s important to consider the cost of running inferences with GPT-[n] and o-[n] models, as these costs can quickly accumulate, especially with more expensive models like o1. Additionally, comparisons between these models and human intelligence should be approached with caution. While we evaluate these models using multimodal puzzles to test their cognitive abilities such as visual perception, inductive reasoning, deductive reasoning, and algorithmic problem-solving, it’s important to recognize that other aspects of human cognition are not being accounted for.

\bibliography{custom}

\appendix

\section{GPT-4o Evaluation Prompt} \label{app:gpt4o_matching}

\begin{tcolorbox}[colback=gray!5, colframe=black, title=GPT-4o Evaluation Prompt]
\footnotesize
\begin{MyVerbatim}
Evaluate the candidate answer against the correct answer. If the candidate answer is correct, output [correct]; otherwise, output [incorrect].

Question: {question}
Candidate Answer: {candidate_answer} 
Correct Answer: {correct_answer}
Evaluation: 
\end{MyVerbatim}
\end{tcolorbox}

\section{\abstractdataset{} Bottleneck Analysis Setup} \label{app:bottleneck}
\begin{figure}[ht]
    \centering
    \includegraphics[width=0.5\linewidth]{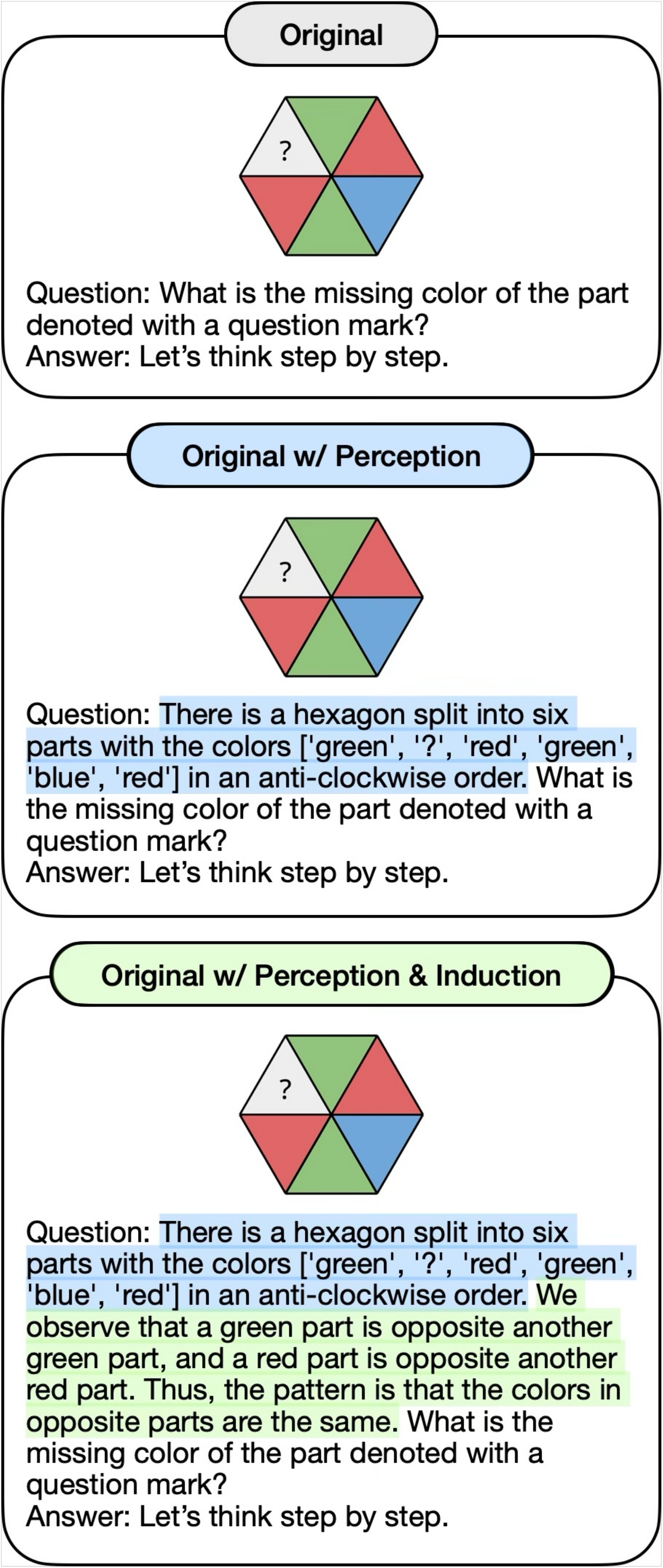}
    \caption{An example of prompts used in the bottleneck analysis: Perception includes the visual details of the puzzle, while induction includes an explanation of the underlying pattern within it.}
    \label{fig:bottleneck}
\end{figure}

\Cref{fig:bottleneck} shows an example of the bottleneck analysis setup.

\section{\abstractdataset{} Case Study Examples} \label{app:abstract_case_study}
\begin{figure*}[ht]
    \centering
    \includegraphics[width=1\linewidth]{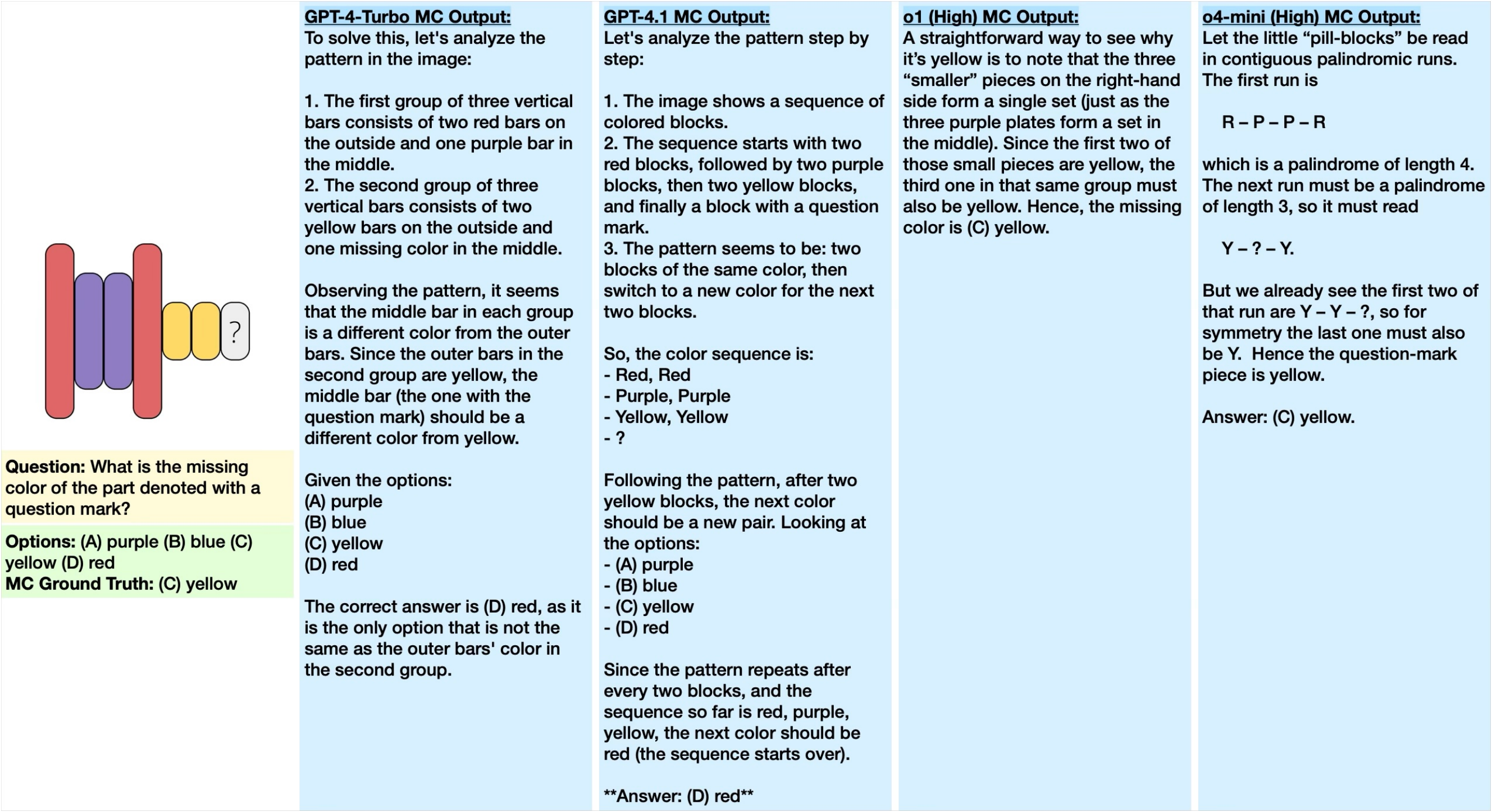}
    \caption{A case study on the \textit{Colors \& Size} puzzle in \abstractdataset{}, evaluated in multiple-choice setting, shows the limitations of GPT-[n] models as compared to o-[n] models.}
    \label{fig:abstract_case_study1}
\end{figure*}

\begin{figure*}[ht]
    \centering
    \includegraphics[width=1\linewidth]{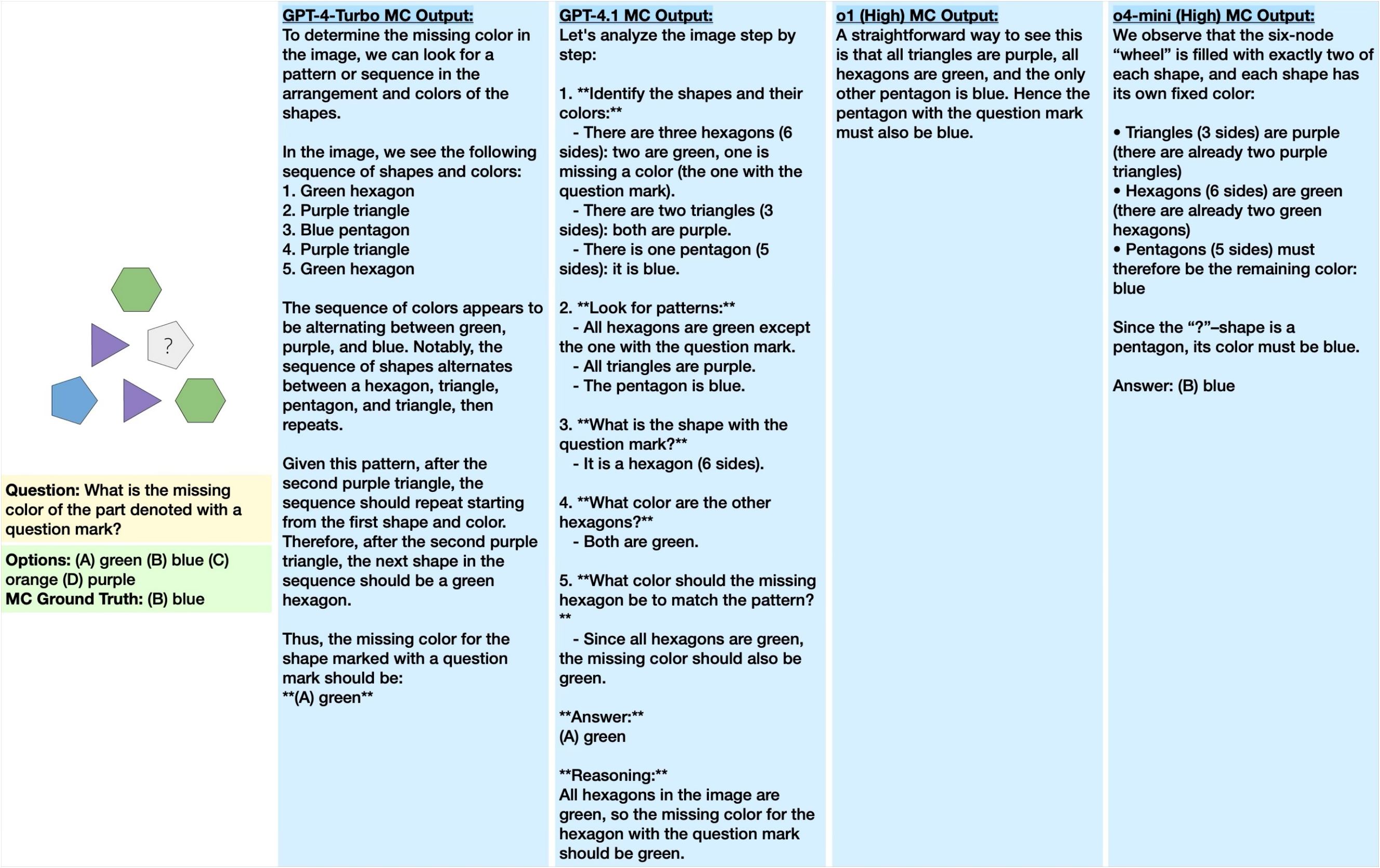}
    \caption{A case study on the \textit{Size \& Shapes} puzzle in \abstractdataset{}, evaluated in multiple-choice setting, shows the limitations of GPT-[n] models as compared to o-[n] models.}
    \label{fig:abstract_case_study2}
\end{figure*}

\begin{figure*}[ht]
    \centering
    \includegraphics[width=1\linewidth]{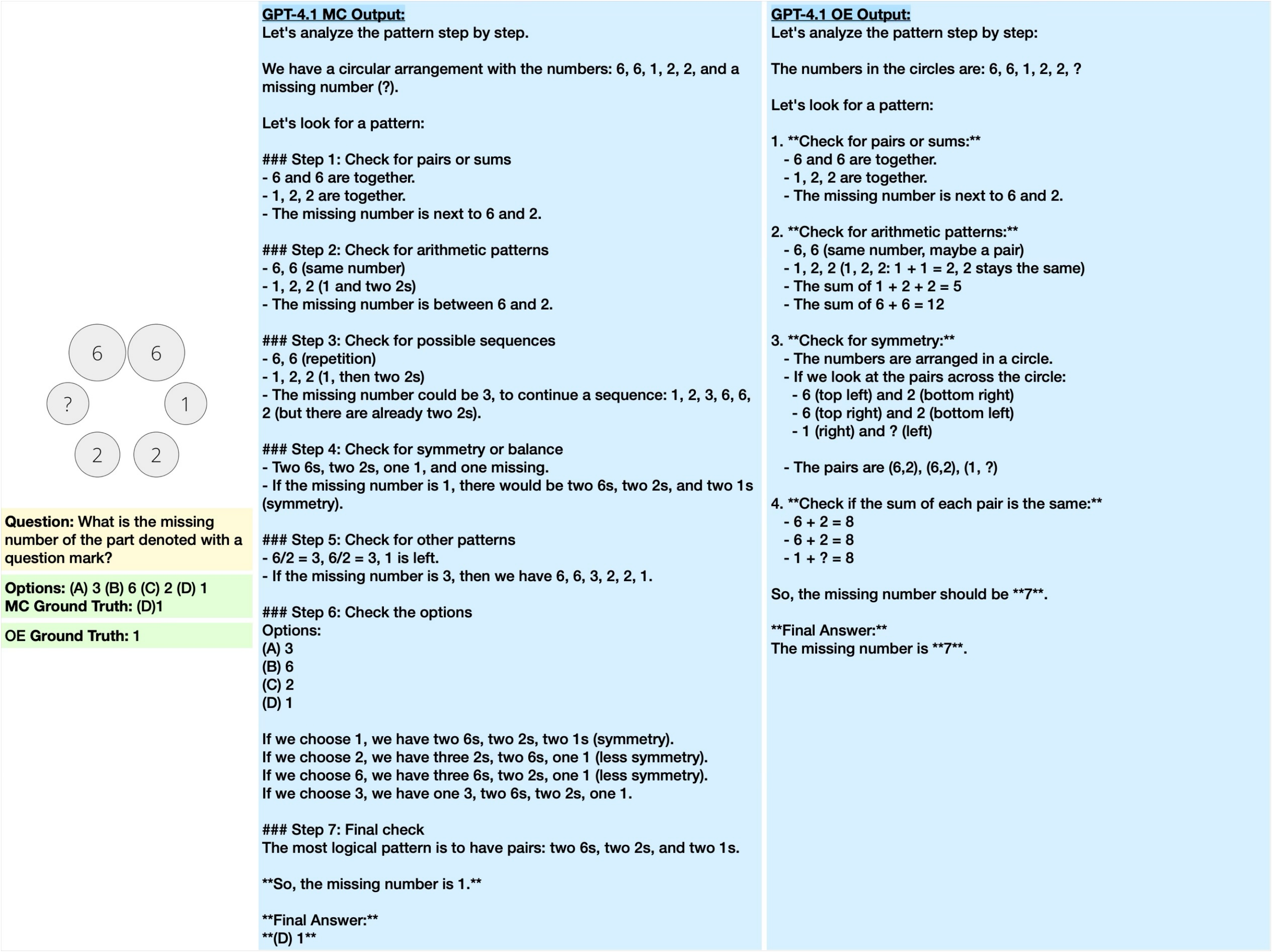}
    \caption{A case study on the \textit{Numbers \& Size} puzzle in \abstractdataset{}, evaluated in both multiple-choice and open-ended settings on GPT-4.1.}
    \label{fig:abstract_case_study3}
\end{figure*}

\begin{figure*}[ht]
    \centering
    \includegraphics[width=1\linewidth]{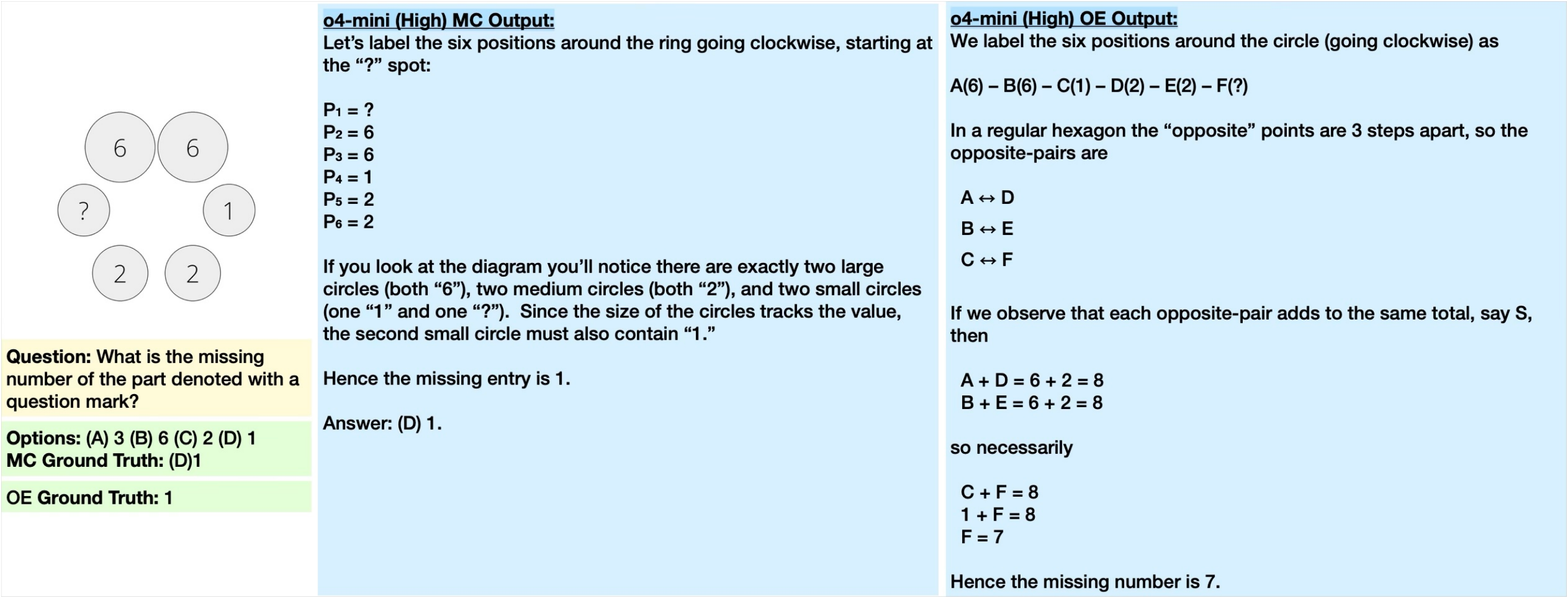}
    \caption{A case study on the \textit{Numbers \& Size} puzzle in \abstractdataset{}, evaluated in both multiple-choice and open-ended settings on o4-mini (High).}
    \label{fig:abstract_case_study4}
\end{figure*}

Figures~\ref{fig:abstract_case_study1} through \ref{fig:abstract_case_study4} present illustrative case studies from the \abstractdataset{} benchmark, highlighting different types of abstract visual reasoning tasks—\textit{Colors \& Size}, \textit{Size \& Shapes}, and \textit{Numbers \& Size}. These examples compare model performance across GPT-[n] and o-[n] variants in both multiple-choice and open-ended formats. The results underscore consistent limitations in the GPT series when faced with compositional and relational reasoning, while showing marked improvements in the o-series, particularly o4-mini, which demonstrates stronger abstraction capabilities across puzzle types.

\section{\algodataset{} Case Study Examples}\label{app:algo_case_study}
\begin{figure*}[ht]
    \centering
    \includegraphics[width=1\linewidth]{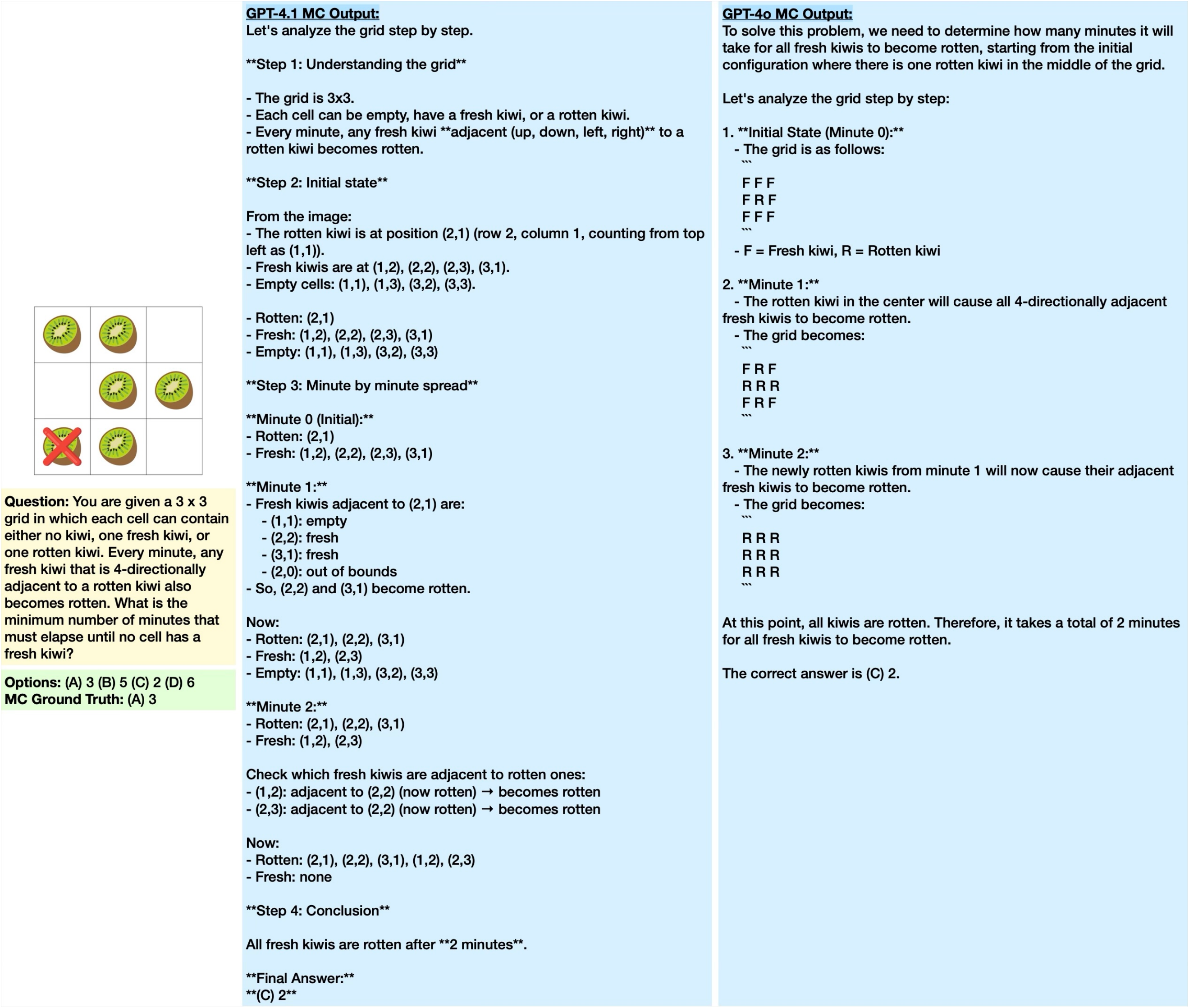}
    \caption{A case study on the \textit{Rotten Fruits} puzzle in \algodataset{}, evaluated in multiple-choice setting on GPT-4.1 and GPT-4o where both models got it wrong.}
    \label{fig:algo_case_study1}
\end{figure*}

\begin{figure*}[ht]
    \centering
    \includegraphics[width=1\linewidth]{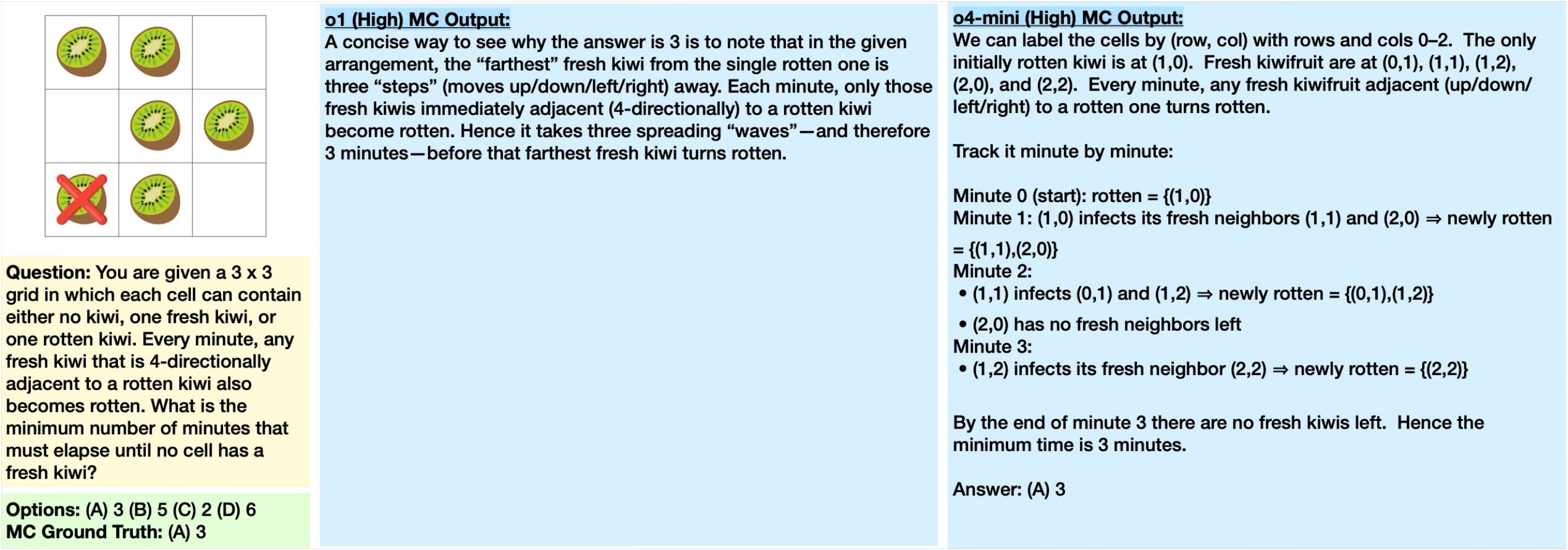}
    \caption{A case study on the \textit{Rotten Fruits} puzzle in \algodataset{}, evaluated in a multiple-choice setting using the o1 (High) and o4-mini (High) models, both of which answered correctly.}
    \label{fig:ealgo_case_study2}
\end{figure*}

\begin{figure*}[ht]
    \centering
    \includegraphics[width=1\linewidth]{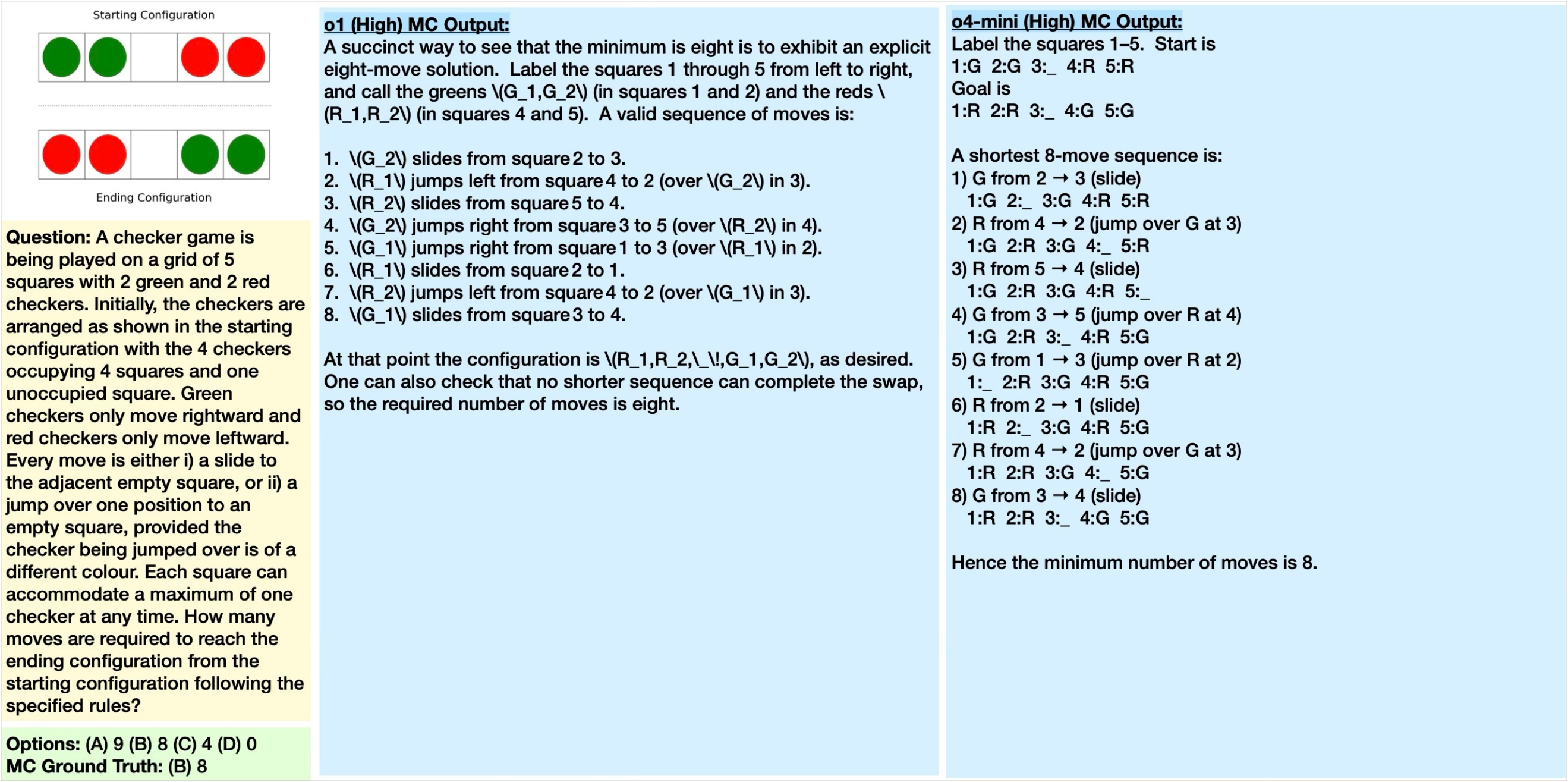}
    \caption{A case study on the \textit{Checker Move} puzzle in \algodataset{}, evaluated in a multiple-choice format using the o1 (High) and o4-mini (High) models, both of which produced correct answers.}
    \label{fig:algo_case_study3}
\end{figure*}

\begin{figure*}[ht]
    \centering
    \includegraphics[width=1\linewidth]{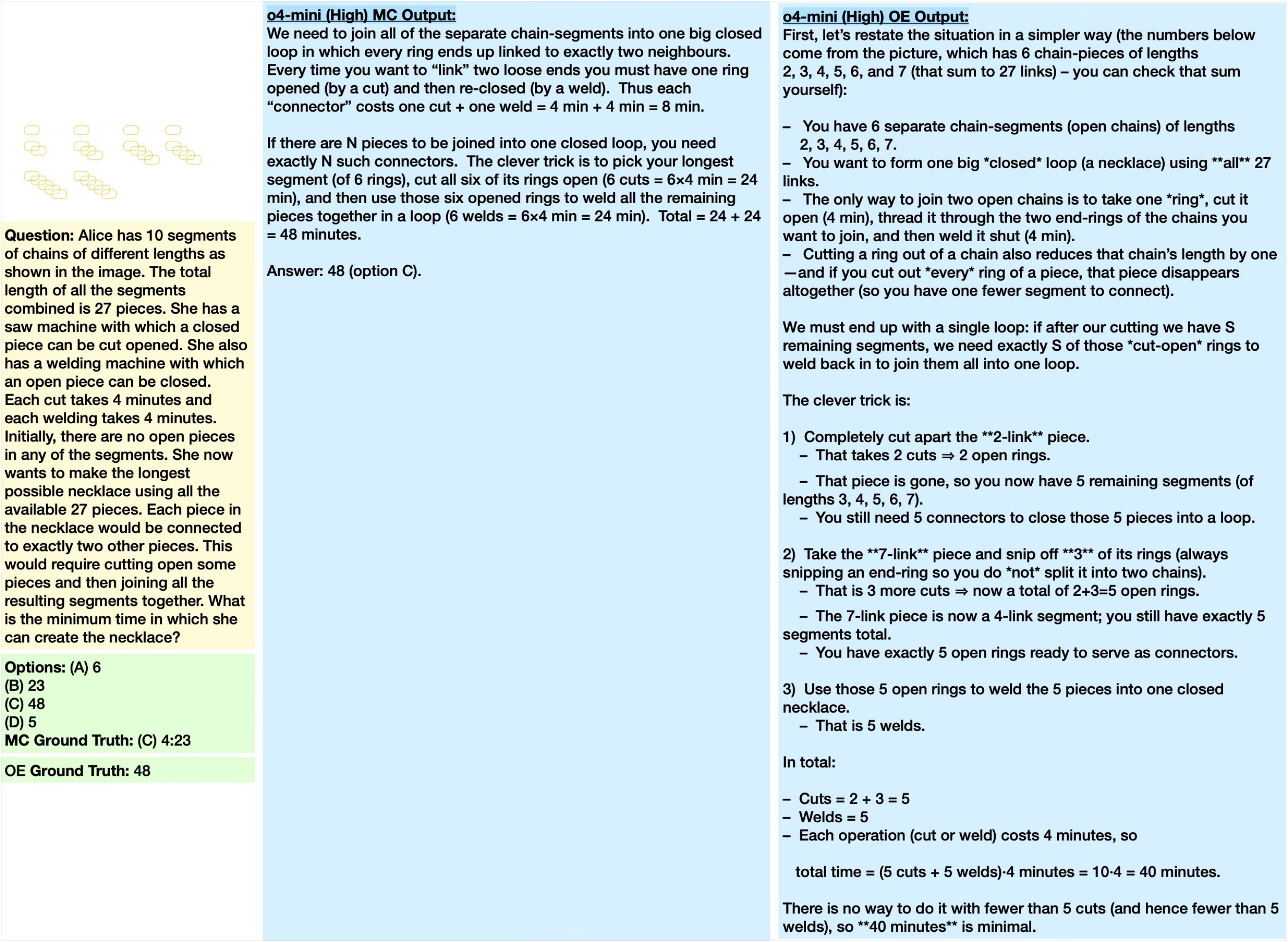}
    \caption{A case study on the \textit{Chain Link} puzzle in \algodataset{}, evaluated in both multiple-choice setting on o4-mini (High) where in the multiple-choice setting the model got it correct and in the open-ended setting, the model got it incorrect}
    \label{fig:algo_case_study4}
\end{figure*}

The case studies in Figures \ref{fig:algo_case_study1}–\ref{fig:algo_case_study4} highlight how model performance on \algodataset{} puzzles varies by model and task format. In the \textit{Rotten Fruits} puzzle, GPT-4.1 and GPT-4o answered incorrectly (Figure~\ref{fig:algo_case_study1}), while o1 (High) and o4-mini (High) answered correctly (Figure~\ref{fig:ealgo_case_study2}), showing improved reasoning in newer models. The same two models also succeeded on the \textit{Checker Move} puzzle (Figure~\ref{fig:algo_case_study3}). In the \textit{Chain Link} puzzle, o4-mini (High) was correct in the multiple-choice setting but failed in the open-ended format (Figure~\ref{fig:algo_case_study4}).

\section{\abstractdataset{} Bottleneck Analysis Case Study Examples}\label{app:bottleneck_case_study}
\begin{figure*}[ht]
    \centering
    \includegraphics[width=1\linewidth]{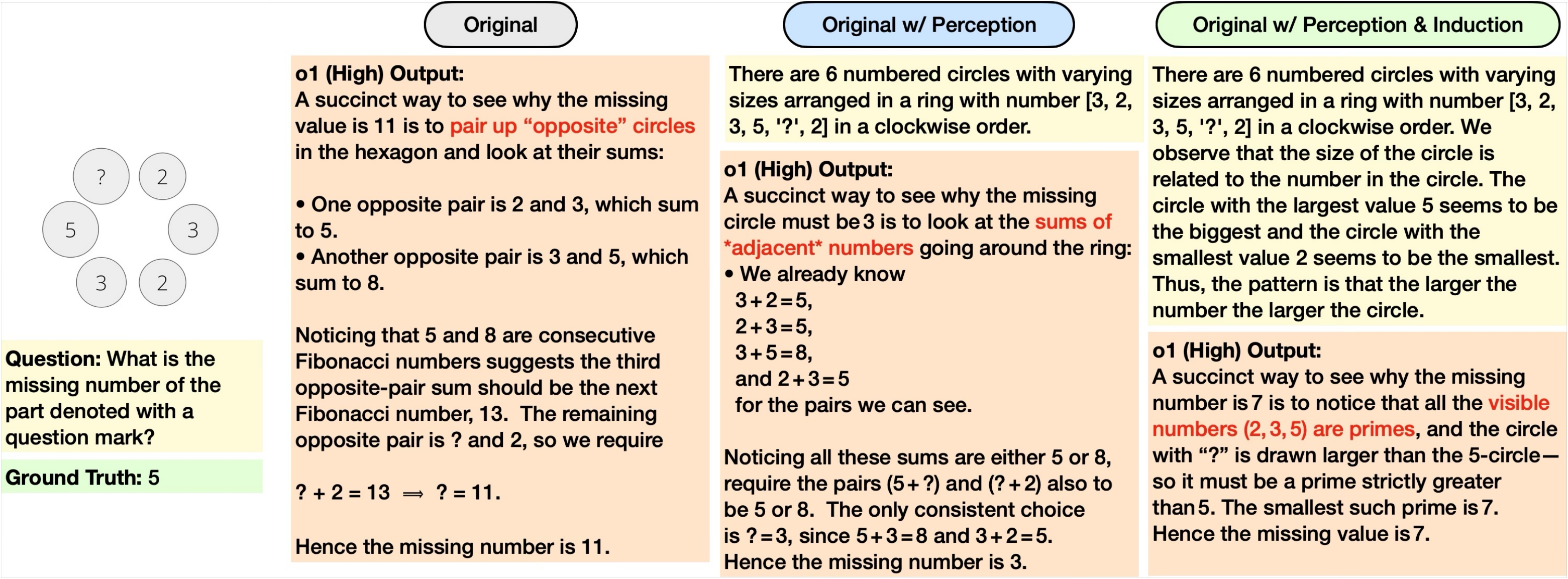}
    \caption{A bottleneck analysis case study on the \textit{Numbers \& Size} puzzle in \abstractdataset{} using o1 (High) reveals that the model failed across all three settings.}
    \label{fig:bottleneck_case_study1}
\end{figure*}

\begin{figure*}[ht]
    \centering
    \includegraphics[width=1\linewidth]{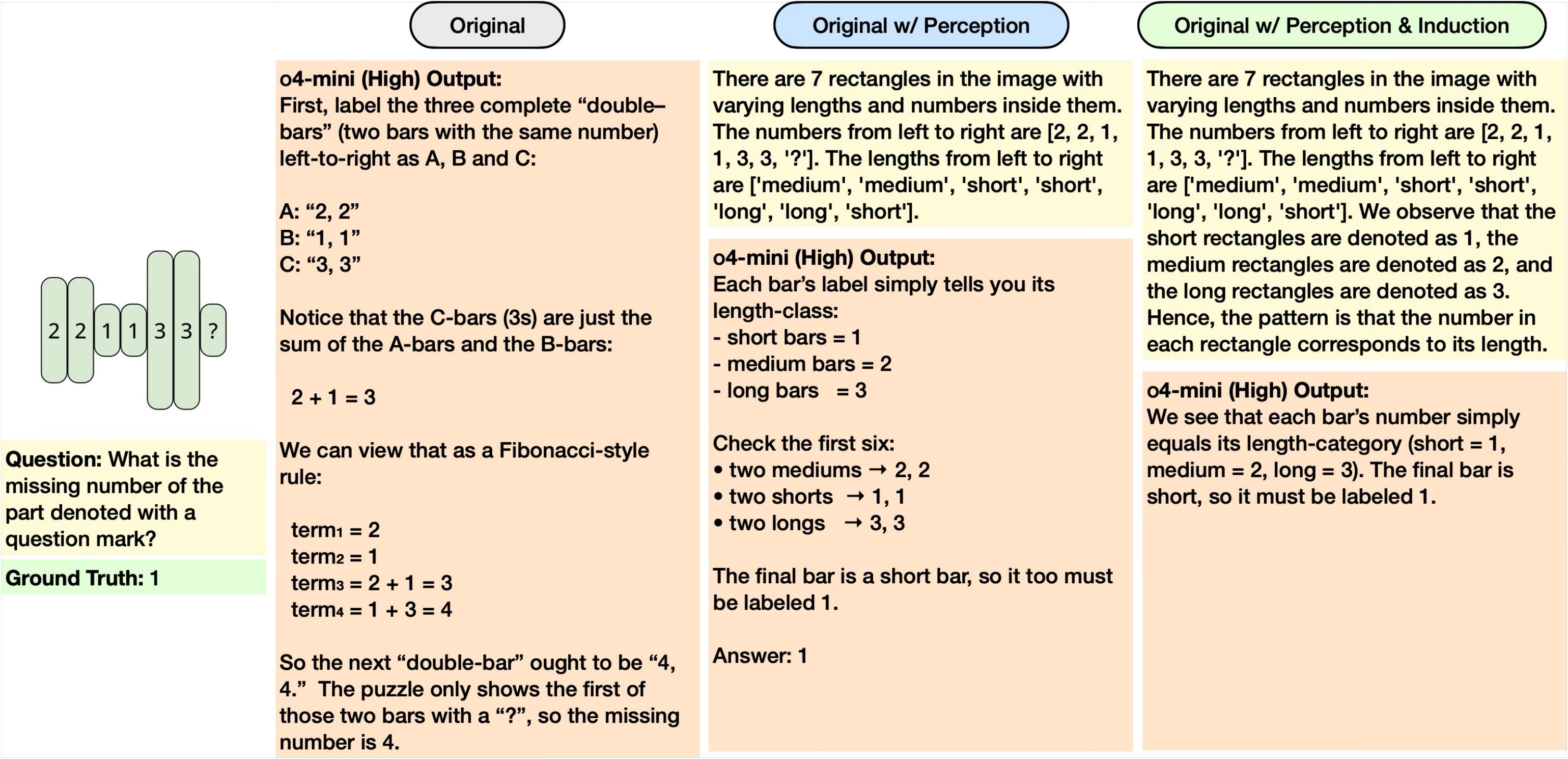}
    \caption{A bottleneck analysis case study on the \textit{Numbers \& Size} puzzle in \abstractdataset{} using o4-mini (High) demonstrates that the model successfully solves the task when provided with visual perception ground truth.}
    \label{fig:bottleneck_case_study2}
\end{figure*}

Figures~\ref{fig:bottleneck_case_study1} and~\ref{fig:bottleneck_case_study2} present bottleneck analysis case studies on the \textit{Numbers \& Size} puzzle from \abstractdataset{}. In Figure~\ref{fig:bottleneck_case_study1}, o1 (High) fails across all conditions, highlighting limitations even with ground-truth visual perception and inductive reasoning. In contrast, Figure~\ref{fig:bottleneck_case_study2} shows that o4-mini (High) succeeds when provided with accurate visual perception input, underscoring the importance of reliable perceptual grounding in solving this task.

\end{document}